\documentclass{article}

    \PassOptionsToPackage{numbers, compress}{natbib}

    \usepackage[final]{neurips_2020}

\usepackage[utf8]{inputenc} 
\usepackage[T1]{fontenc}    
\usepackage{hyperref}       
\usepackage{url}            
\usepackage{booktabs}       
\usepackage{amsfonts}       
\usepackage{nicefrac}       
\usepackage{microtype}      
\usepackage{amsmath}
\usepackage{natbib}
\usepackage{algorithm}
\usepackage{booktabs} 
\usepackage{caption}
\usepackage{xcolor}
\usepackage[noend]{algpseudocode}
\usepackage{wrapfig}
\usepackage{adjustbox}
\usepackage{subfigure}
\usepackage{changepage}
\usepackage{sidecap}

\def\eg{\emph{e.g.~}}

\def\ie{\emph{i.e.~}}
\def\aka{{\em a.k.a.~}}

\title{Exposing backdoor attacks with the help of adversarial examples
}

\title{Adversarial examples are useful too!
}

\author{
Ali Borji \\
\texttt{aliborji@gmail.com} 
\thanks{Code is available at: \url{https://github.com/aliborji/Backdoor_defense.git}}
}

\begin{document}

\maketitle

\begin{abstract}
    Deep learning has come a long way and has enjoyed an unprecedented success. Despite high accuracy, however, deep models are brittle and are easily fooled by imperceptible adversarial perturbations. In contrast to common inference-time attacks, Backdoor (\aka Trojan) attacks target the training phase of model construction, and are extremely difficult to combat since a) the model behaves normally on a pristine testing set and b) the augmented perturbations can be minute and may only affect few training samples. Here, I propose a new method to tell whether a model has been subject to a backdoor attack. The idea is to generate adversarial examples, targeted or untargeted, using conventional attacks such as FGSM and then feed them back to the classifier. By computing the statistics (here simply mean maps) of the images in different categories and comparing them with the statistics of a reference model, it is possible to visually locate the perturbed regions and unveil the attack. 
 \end{abstract}



\section{Introduction}
\begin{figure}[htbp]       
\centering
\vspace{-5pt}
\includegraphics[width=.22\linewidth]{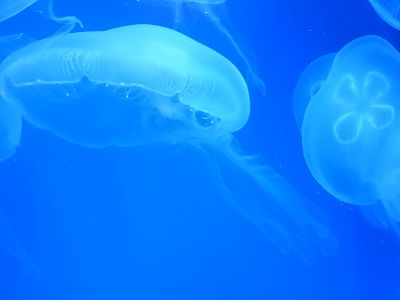}
\includegraphics[width=.22\linewidth]{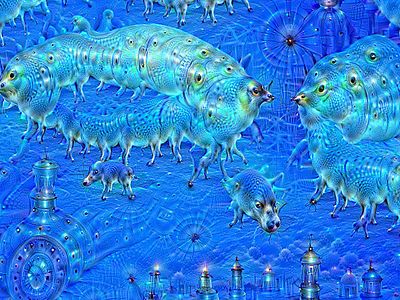}
\includegraphics[width=.22\linewidth]{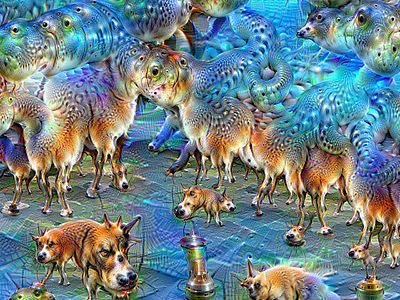}
\caption{An example image (left) after ten (middle) and fifty (right) iterations of DeepDream, using a network trained to perceive dogs. Figure from \url{https://en.wikipedia.org/wiki/DeepDream}.}
\label{fig:deepdream}
\end{figure}

Adversarial examples are crafted to fool deep learning models~\cite{szegedy2013intriguing,goodfellow2015explaining}, and by that virtue they are deemed undesirable. How can they be useful then? Well, they somehow have been indirectly utilized in the past. Propagating the loss with respect to the input, instead of the weights, has been used in several related but seemingly different applications including DeepDream\footnote{\url{https://ai.googleblog.com/2015/07/deepdream-code-example-for-visualizing.html}} (Fig.~\ref{fig:deepdream}), Style transfer~\cite{gatys2016image}, feature visualization~\cite{zeiler2014visualizing,Selvaraju_2017_ICCV}, and adversarial attacks~\cite{goodfellow2015explaining}. Here, I will show how this technique can be used to unveil interference in a classifier by means of poisoning attacks~\cite{liu2016delving}. In other words, I am proposing a simple defense against backdoor poisoning attacks. I will provide minimal experiments to prove the concept. Further experiments, however, are needed to scale up this approach over more complex datasets.


Two types of adversarial attacks are common against deep learning models, \emph{evasion} and \emph{poisoning}. In evasion attacks, the adversary applies an imperceptible perturbation, digital or physical, to the input image in order to fool the classifier (\eg~\cite{szegedy2013intriguing,goodfellow2015explaining}). The perturbation can be untargeted (changing the decision to any output other than the true class label) or targeted (changing the decision to a class of interest), black-box (when only available information is the output labels or scores) or white box (some information such as model architecture or gradients are available). The goal in poisoning attacks (\eg~\cite{gu2017badnets,liu2017trojaning,zhang2016rethinking,wang2019neural,brown2017adversarial}) is to either a) introduce infected samples with wrong labels to the training set to degrade the testing accuracy of the model (\aka collision attacks) or b) introduce a trigger (\eg a sticker) during training such that activating the trigger at testing time will initiate a malfunction in the system (\aka backdoor attacks). Here, I am concerned with backdoor poisoning attacks.


The problem statement is as follows. Given a model, and possibly a reference model that is supposed to be used solely for fine tuning (\eg a backbone), is it possible to tell whether the model contains a backdoor attack? The approach I take here is based on the idea that adversarial examples generated for an attacked model would be different than those generated for a clean reference model. In what follows I will explain how we can tap into such differences to discover anomalies.


\begin{figure}[t]       
    \centering
    
    \fbox{\includegraphics[width=\linewidth]{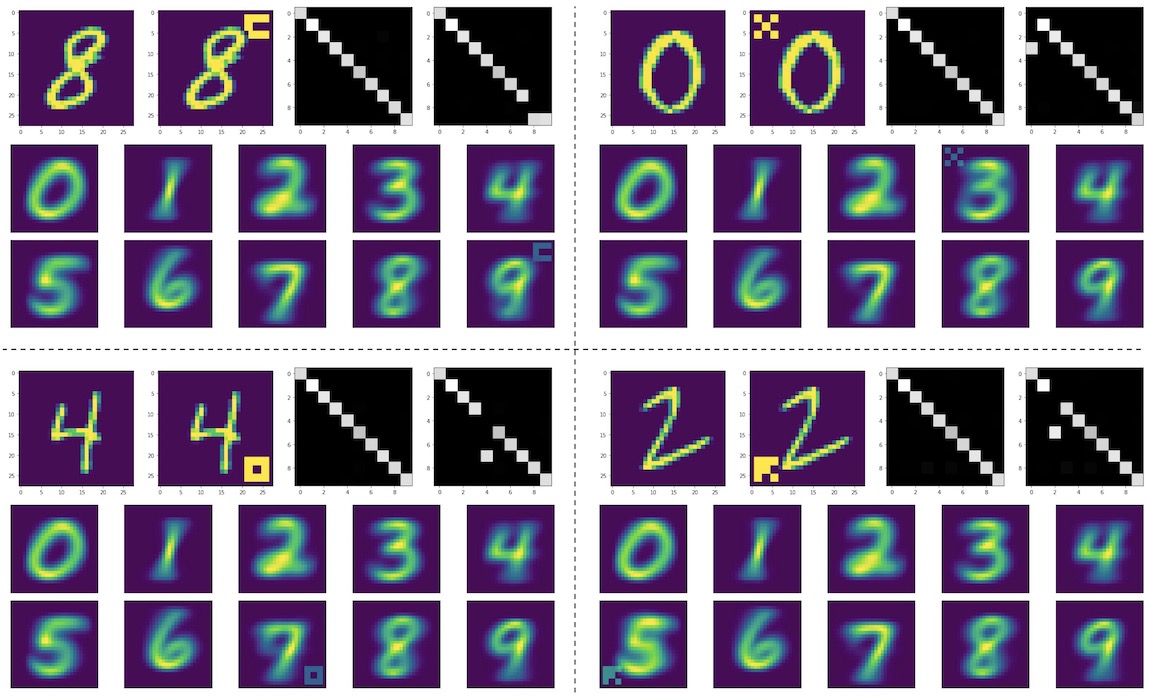}} 
    \caption{Each of the four panels represents a backdoor poisoning attack over MNIST. A sample image and its corresponding patch-augmented image, confusion matrices over the pristine testing dataset and the poisoned dataset (a patch added to all digits of one class), as well as the average digit maps are shown.}
    \label{fig:fig0}
\end{figure}

\section{Approach}
To diagnose the attack, first we need to generate adversarial examples. We can feed three types of inputs to the model a) Blank image, b) White noise, c) (Unlabeled) Data, and then use an inference-time adversarial attack method (here FGSM~\cite{goodfellow2015explaining}) to craft adversarial examples:
\begin{equation}
\mathbf{x}_{adv} \leftarrow \text{clip}(\mathbf{x} - \epsilon \times sign(\nabla_{\mathbf{x}} J(\mathbf{\theta}, \mathbf{x}, y_{target})))
\end{equation}
Here, $\mathbf{x}$ is the input image and $y_{target}$ is the target class we wish the input to be classified as, and $\epsilon \in [0,1]$ controls the magnitude of perturbation. Notice that the attack does not necessarily need to be targeted. In practice, I apply the FGSM for a number of iterations akin to IFGSM~\cite{kurakin2016physical}. Then, the average of all adversarial examples that are classified as a certain class is computed (Algorithm I). The same procedure is repeated for the clean reference model. As will be shown later, over simple datasets such as MNIST, visual inspection of the average adversarial images (also called bias maps) is enough to discover the attack. Over more complex datasets such as CIFAR, however, more sophisticated quantitative methods such as outlier detection might be necessary.

\begin{verbatim}
Algorithm I:
for target in categories:
    for a number of samples:
        generate x_adv according to Eq. 1
        prediction <-- model(x_{adv})
        record x_{adv} under prediction
    end for        
end for
\end{verbatim}

While the above method is effective, it is not granular enough to reveal all categories that are impacted. To address this, we can perturb images such that they are more likely to belong to the target category and less likely to belong to the source\footnote{Do not confuse this with the category in which images are perturbed.} category:
\begin{equation}
\mathbf{x}_{adv} \leftarrow \text{clip}(\mathbf{x} - \epsilon \times sign(\nabla_{\mathbf{x}} (J(\mathbf{\theta}, \mathbf{x}, y_{target}) - J(\mathbf{\theta}, \mathbf{x}, y_{source}))))
\end{equation}

For each pair of categories (except when \emph{source=target}), a number of adversarial examples are generated. The average of adversarial images for each pair is then computed. To be a bit more precise:
\begin{verbatim}
Algorithm II:
for source in categories:
    for target in categories:
        if source == target: continue
        for a number of samples:
            generate x_adv according to Eq. 2
            prediction <-- model(x_{adv})
            record x_{adv} under (source, prediction)
        end for
    end for
end for
\end{verbatim}

\section{Experiments and Results}
Experiments are conducted over the MNIST dataset. In each experiment, first a CNN (Fig.~\ref{fig:model}) is trained over the pristine data (called clean model in Figures). Then, an adversarial patch is added to 2000 randomly selected images from the source category and their labels are changed to the target category (\eg $8\rightarrow9)$. The remaining images from the source category preserve their labels. The CNN is then fine-tuned on these data (called backdoor model in Figures). Please see Fig.~\ref{fig:fig0}.

\subsection{Experiment I}
Adversarial backdoor attacks were planted over four different source-target pairs: $8\rightarrow9$, $0\rightarrow3$, $4\rightarrow7$, and $2\rightarrow5$. Sample images, augmented images with the adversarial patches, confusion matrices of the backdoor models on pristine and poisoned testing sets, as well as mean digit images are shown in Fig.~\ref{fig:fig0}. Both clean and backdoor models perform very well above 95\%. Applying the backdoor model to the pristine testing sets everything looks normal and confusion matrices resemble those of the clean model. However, when evaluated on a testing set with all source images poisoned and relabeled with the target category, almost all of those images are now classified as the target class (the top right matrix in each panel of Fig.~\ref{fig:fig0}). 

\begin{figure}[htbp]       
\begin{adjustwidth}{-2cm}{-2cm}
\includegraphics[width=1\linewidth]{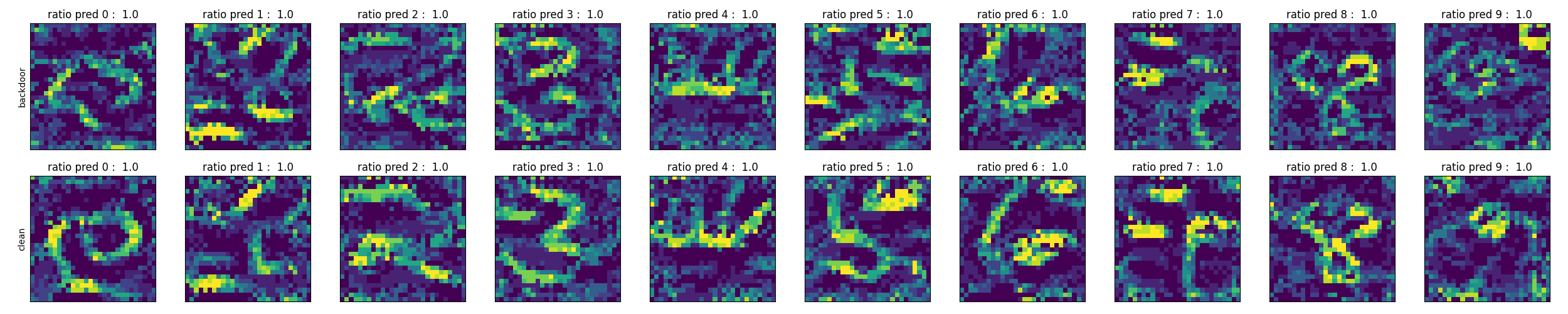}
\includegraphics[width=1\linewidth]{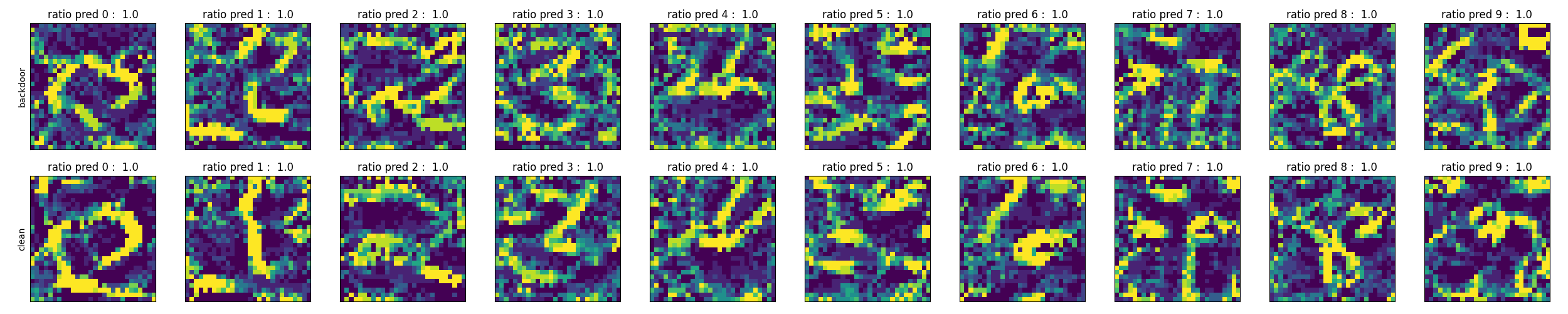}
\vspace{25pt} 
\includegraphics[width=1\linewidth]{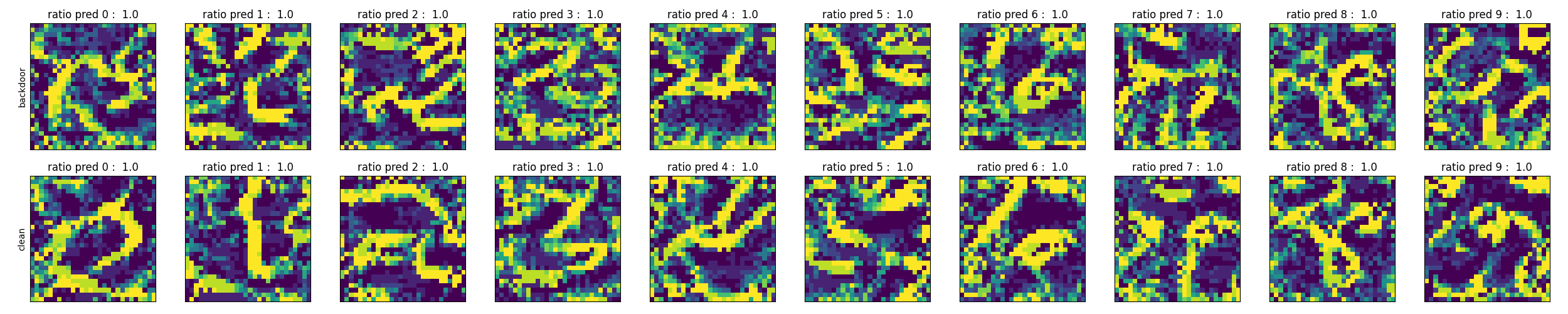}
\includegraphics[width=1\linewidth]{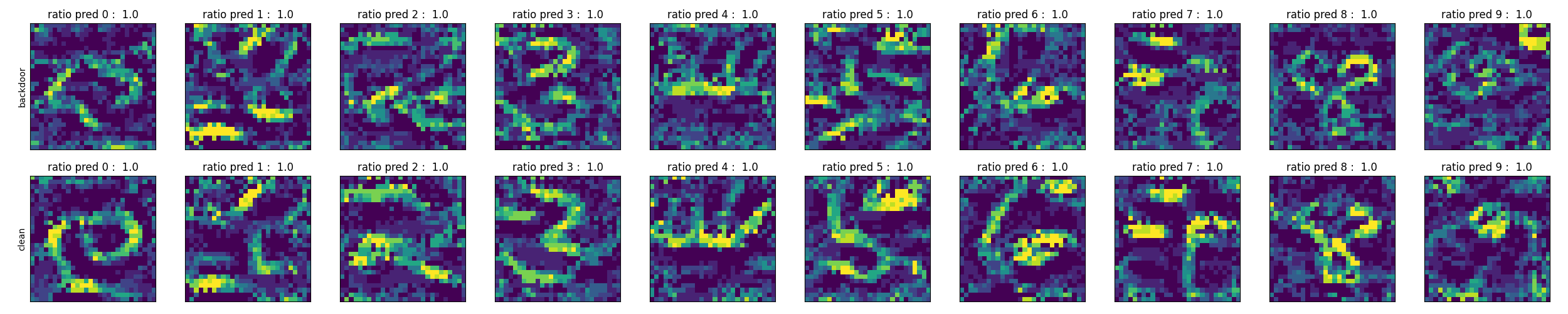}
\caption{Bias maps derived using blank inputs to the Algorithm I for the $8\rightarrow9$ attack. Plots are in pairs corresponding to the attacked and clean models, respectively. 
Results are shown for 10 (first two rows), 30 (second two rows), and 100 iterations of the FGSM attack using one blank image and $\epsilon$=0.1. 
The bottom rows show the results with 100 samples, 10 iterations and $\epsilon$=0.1. Increasing the number of samples does change the outcome. Image titles show the percentage of the images that are classified under a particular class. Notice, however, that bias maps are all computed for the images that fall under a certain category.
}
\label{fig:89-blank}
\end{adjustwidth}
\end{figure}

Fig.~\ref{fig:89-blank} shows the results of applying algorithm I to the models for the $8\rightarrow9$ attack using blank images as input to FGSM. A single image is iteratively modified to be categorized as the target class. Two observations can be made here. First, bias maps resemble the digits, and second, the trigger at the top-right corner of the bias map of category 9 is highlighted. The same region is off for the clean model. The bias maps of other categories are similar across clean and attacked models. Expectedly, I found that bias maps for the clean models do not change much across the experiments. I also found that moderate values for $\epsilon$ leads to better results.

\begin{figure}[htbp]     
\vspace{-20pt}
\begin{adjustwidth}{-2cm}{-2cm}
\ \ \ \ Using data \\
\includegraphics[width=1\linewidth]{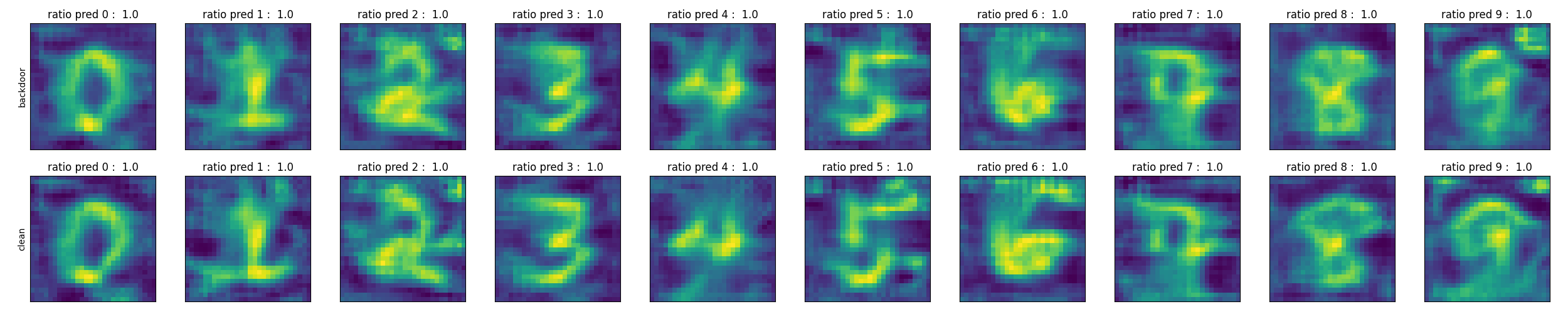}
\includegraphics[width=1\linewidth]{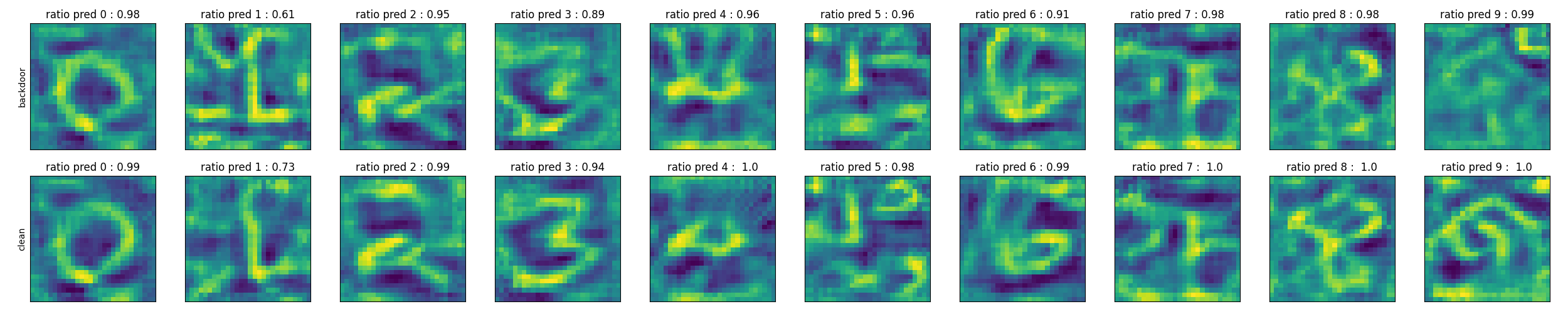}
\vspace{15pt}
\includegraphics[width=1\linewidth]{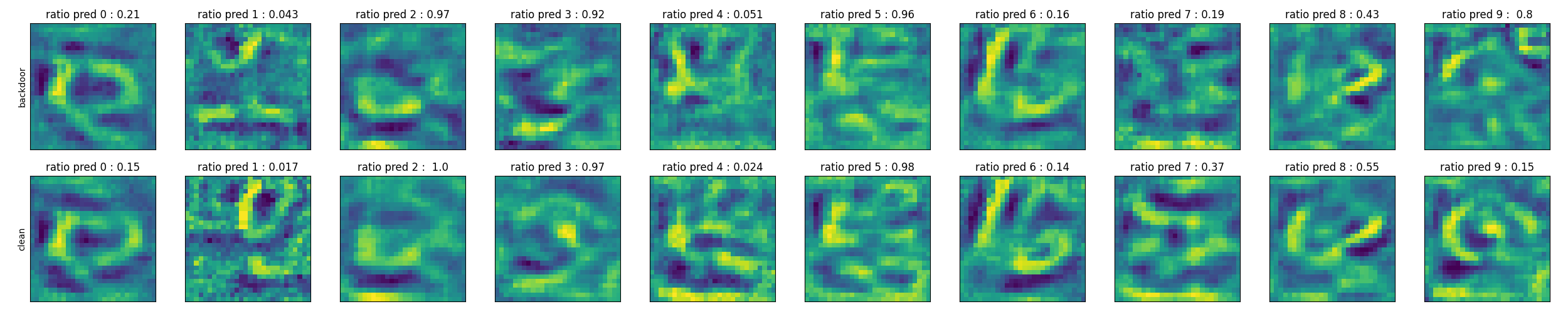}

\ \ \ \ Using noise \\
\includegraphics[width=1\linewidth]{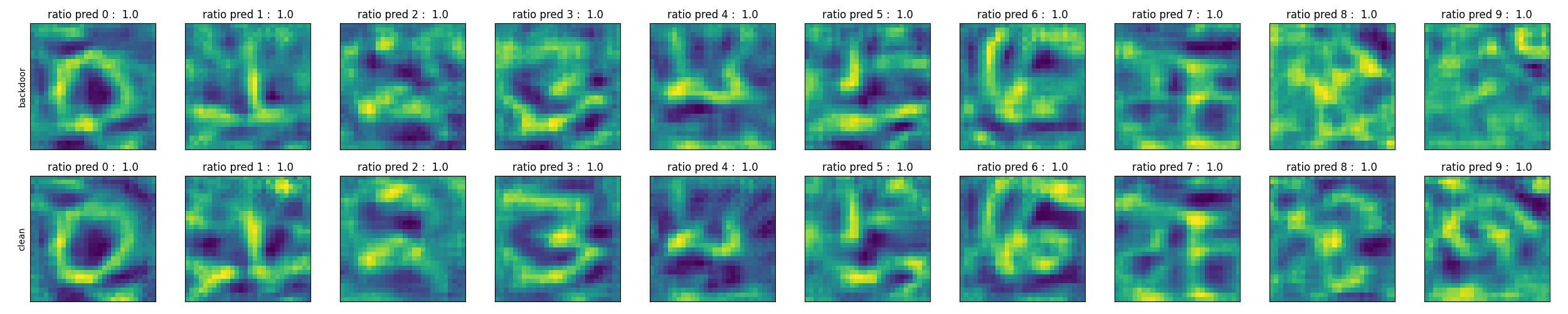}
\includegraphics[width=1\linewidth]{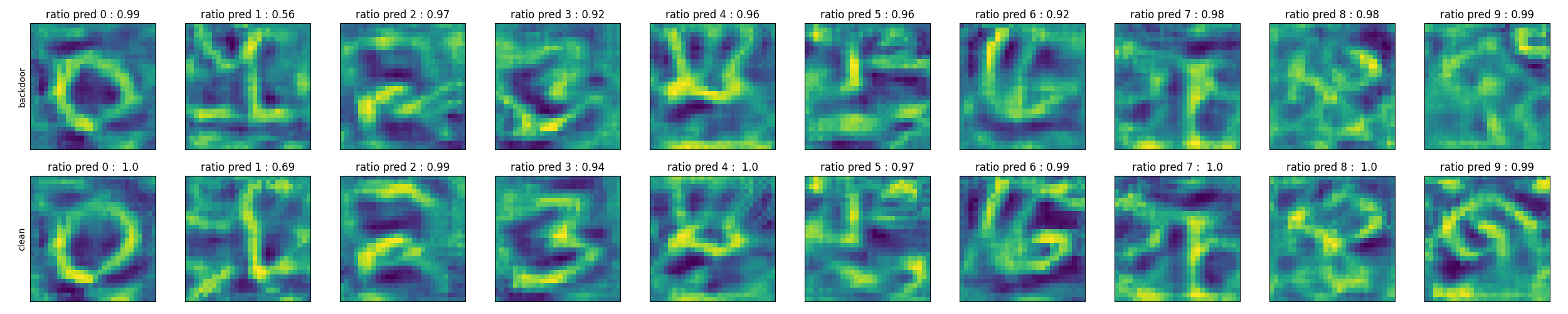}
\vspace{15pt}
\includegraphics[width=1\linewidth]{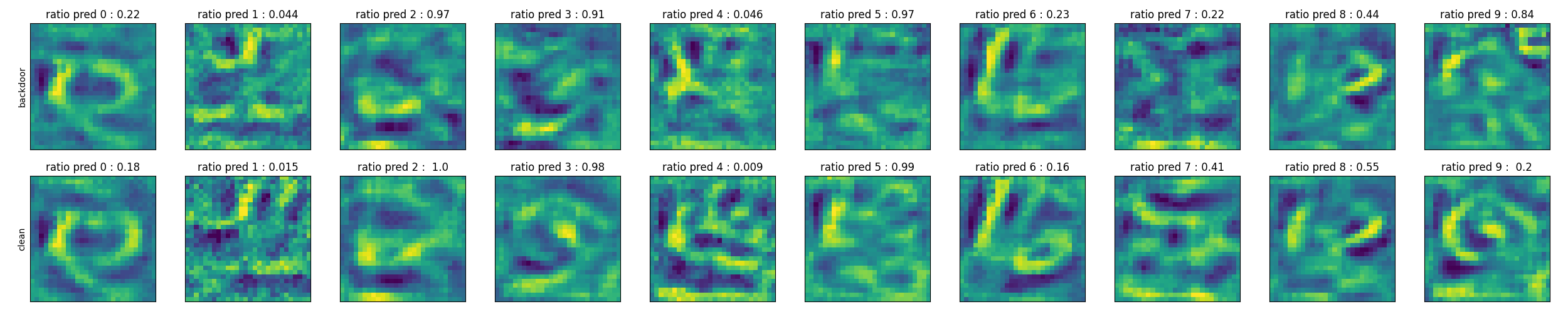}
\vspace{-30pt}
\caption{Bias maps derived by feeding data (top) and white noise (bottom) to Algorithm I, for the $8 \rightarrow 9$ attack. In each panel, pairs from top to bottom correspond to $\epsilon$ equal to 0.1, 0.5, and 0.9, respectively. Number of samples and iterations were set to 1000, 10, respectively.}
\label{fig:89}
\end{adjustwidth}
\end{figure}

\begin{figure}[htbp]       
\vspace{-20pt}
\begin{adjustwidth}{-2cm}{-2cm}
\ \ \ \ Using data \\
\includegraphics[width=1\linewidth]{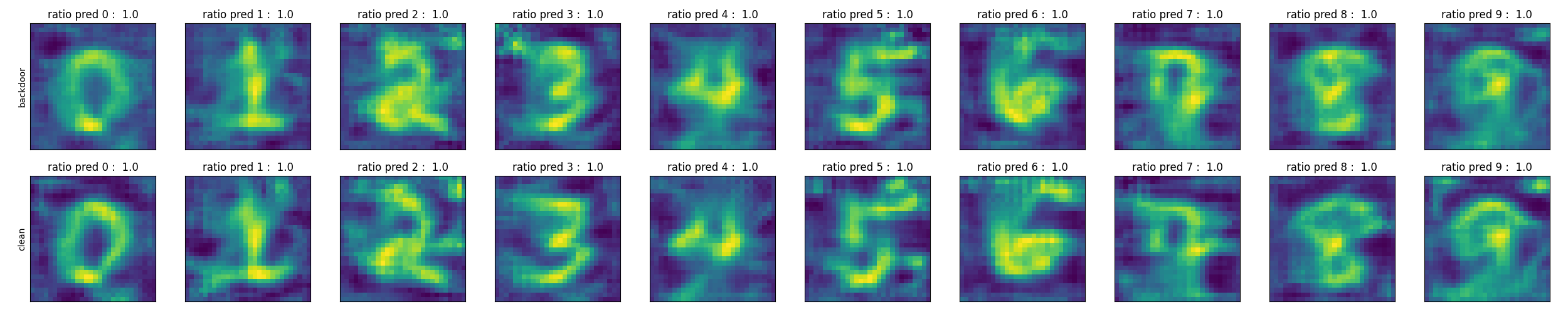}
\includegraphics[width=1\linewidth]{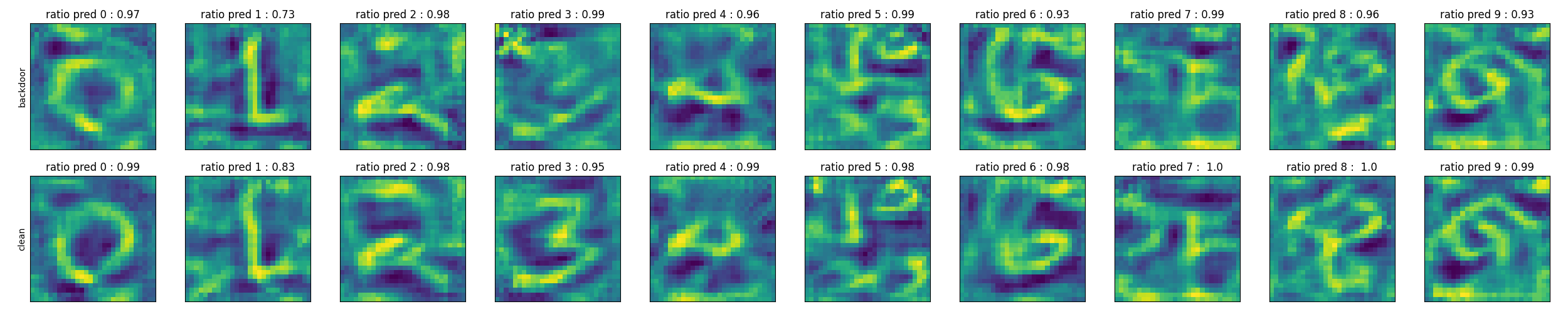}
\vspace{15pt}
\includegraphics[width=1\linewidth]{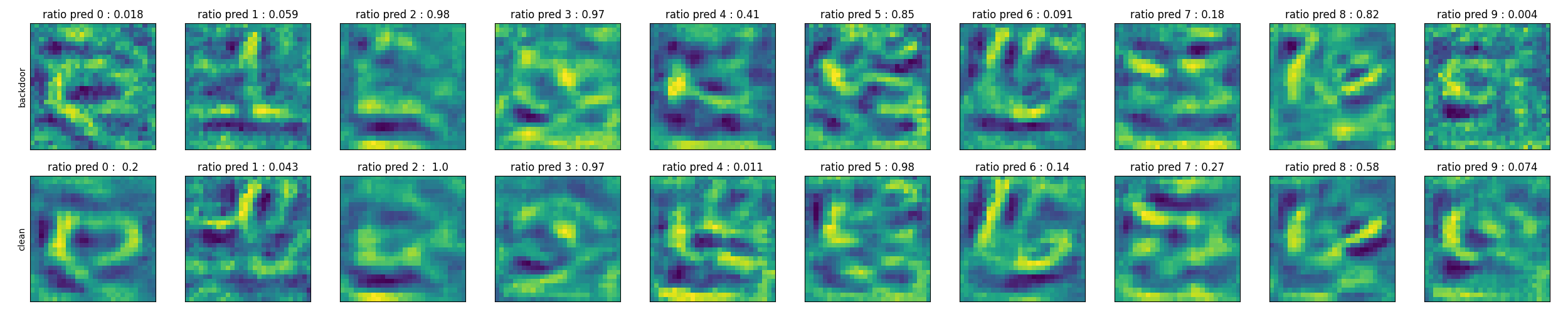}

\ \ \ \ Using noise \\
\includegraphics[width=1\linewidth]{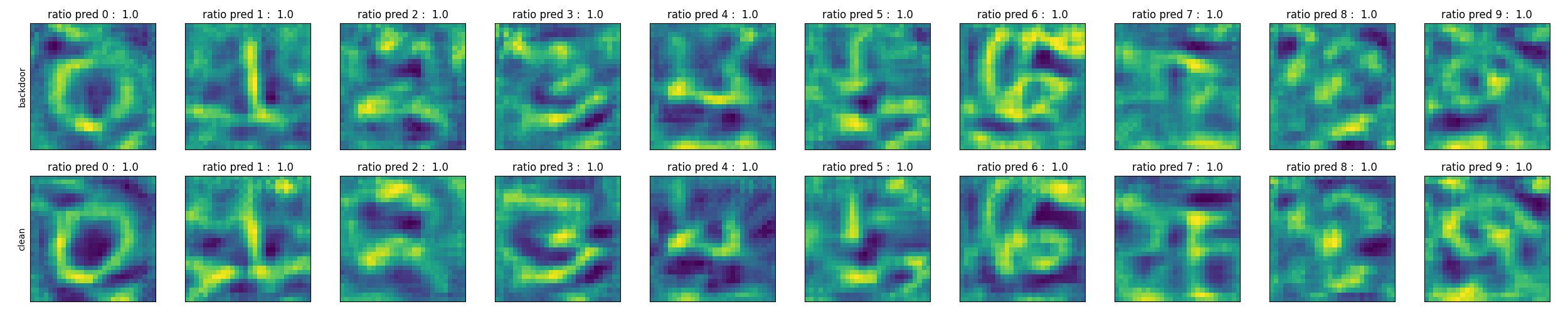}
\includegraphics[width=1\linewidth]{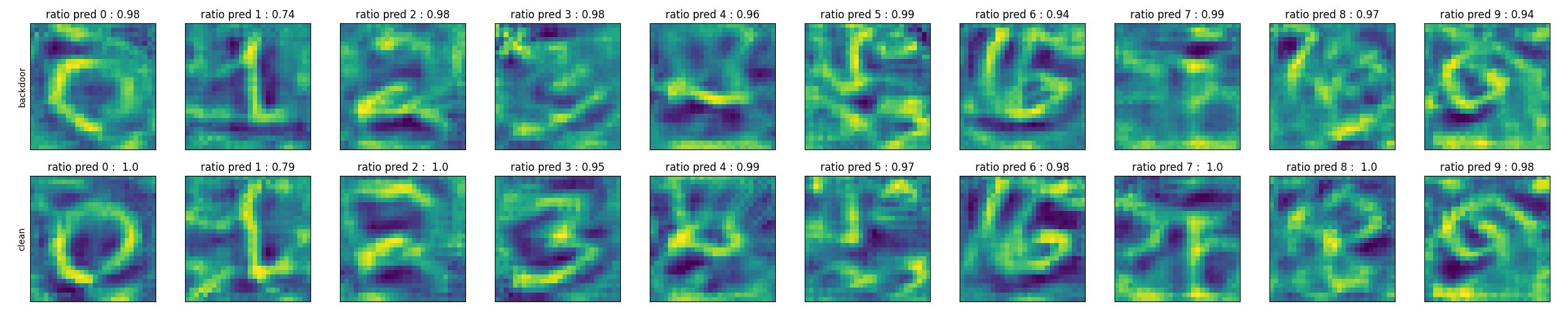}
\vspace{15pt}
\includegraphics[width=1\linewidth]{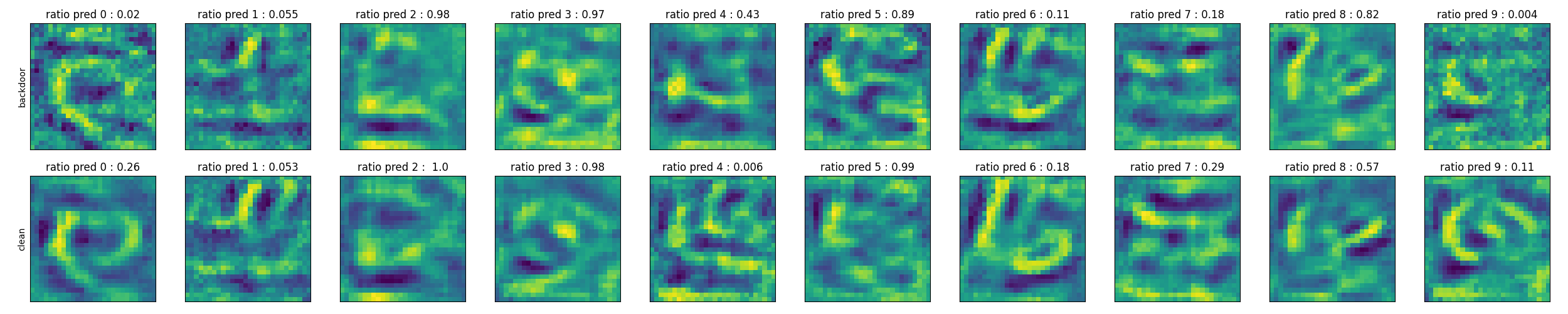}
\vspace{-30pt}
\caption{Bias maps derived by feeding data (top) and white noise (bottom) to Algorithm I, for the $0 \rightarrow 3$ attack. In each panel, pairs from top to bottom correspond to $\epsilon$ equal to 0.1, 0.5, and 0.9, respectively. Number of samples and iterations were set to 1000, 10, respectively.}
\label{fig:03}
\end{adjustwidth}
\end{figure}

\begin{figure}[htbp]   
\vspace{-20pt}
\begin{adjustwidth}{-2cm}{-2cm}
\ \ \ \ Using data \\
\includegraphics[width=1\linewidth]{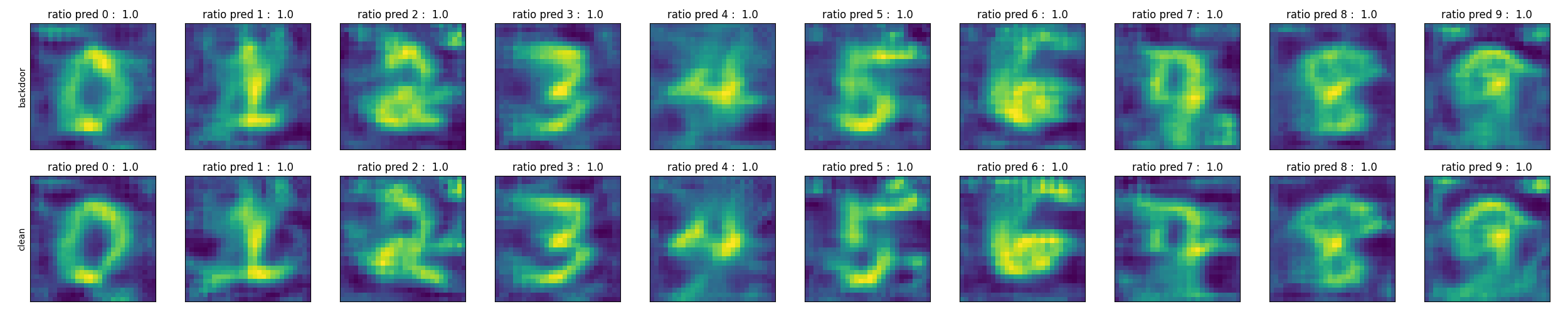}
\includegraphics[width=1\linewidth]{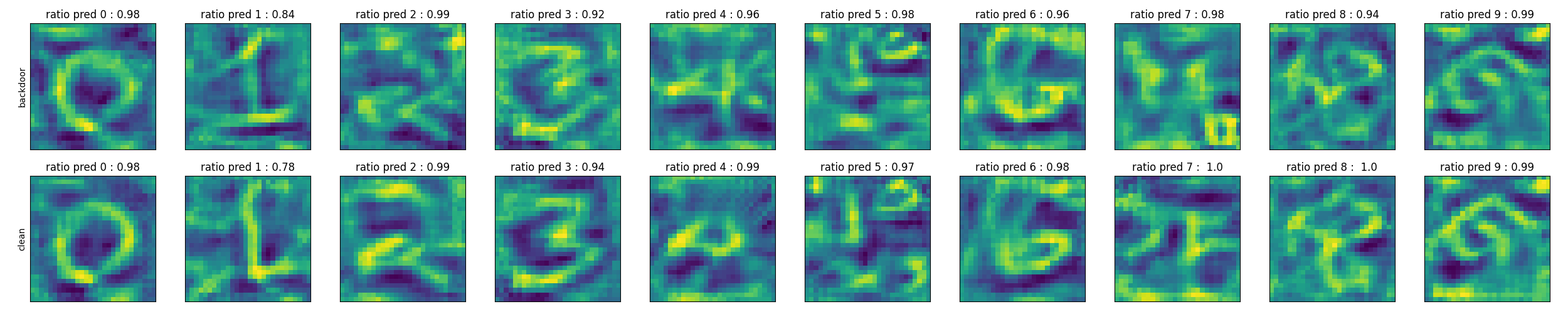}
\vspace{15pt}
\includegraphics[width=1\linewidth]{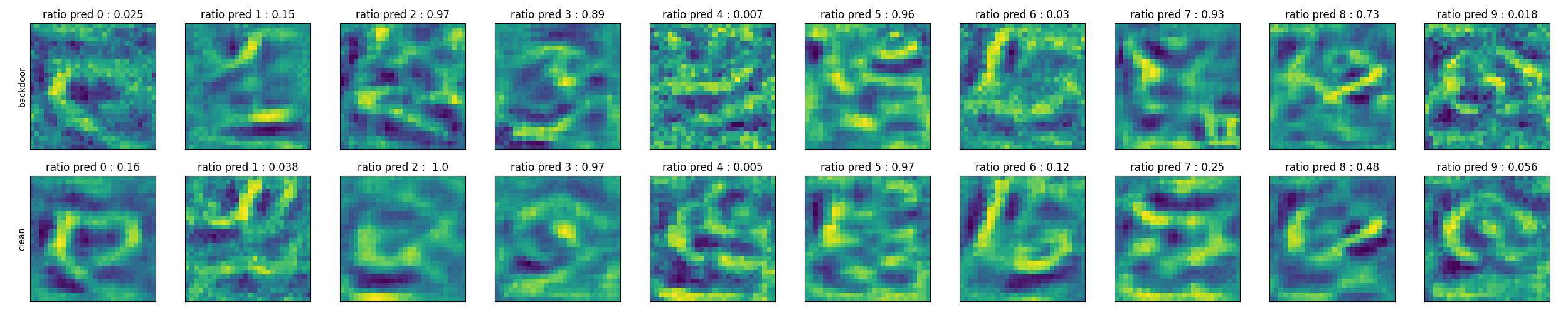}

\ \ \ \ Using noise \\
\includegraphics[width=1\linewidth]{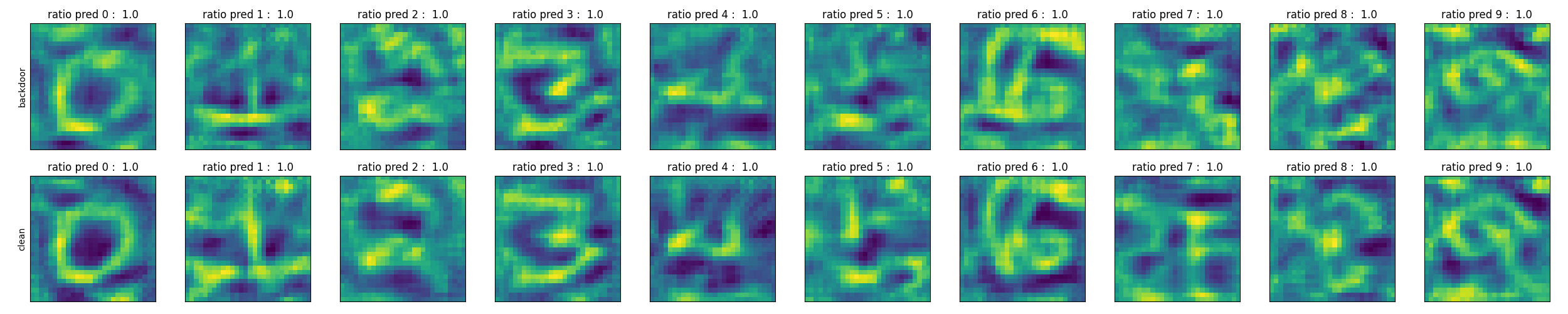}
\includegraphics[width=1\linewidth]{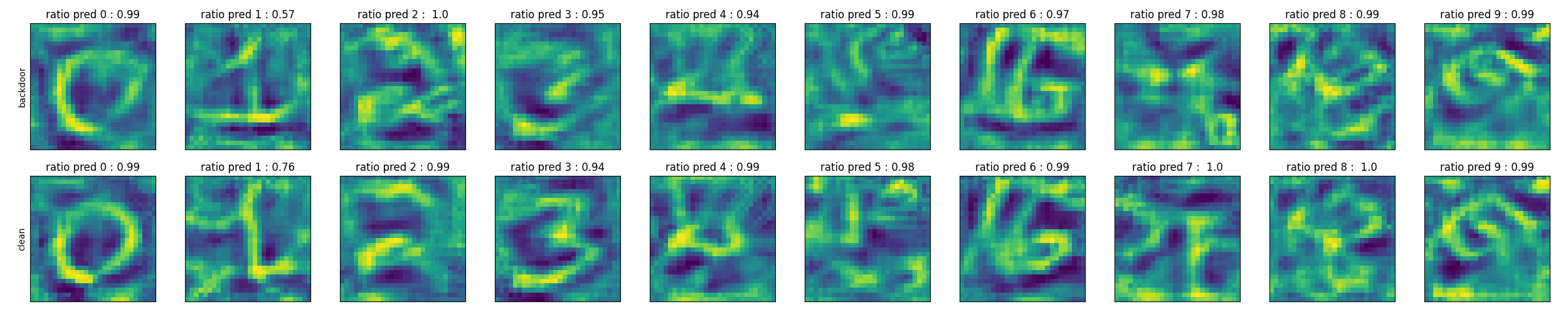}
\vspace{15pt}
\includegraphics[width=1\linewidth]{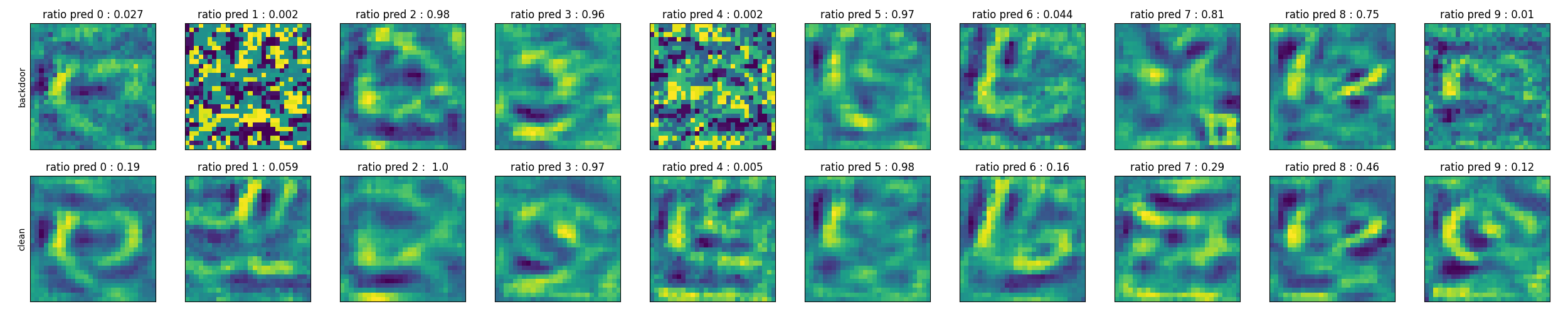}
\vspace{-30pt}
\caption{Bias maps derived by feeding data (top) and white noise (bottom) to Algorithm I, for the $4 \rightarrow 7$ attack. In each panel, pairs from top to bottom correspond to $\epsilon$ equal to 0.1, 0.5, and 0.9, respectively. Number of samples and iterations were set to 1000, 10, respectively.}\label{fig:47}
\end{adjustwidth}
\end{figure}

\begin{figure}[htbp]       
\vspace{-20pt}
\begin{adjustwidth}{-2cm}{-2cm}
\ \ \ \ Using data \\
\includegraphics[width=1\linewidth]{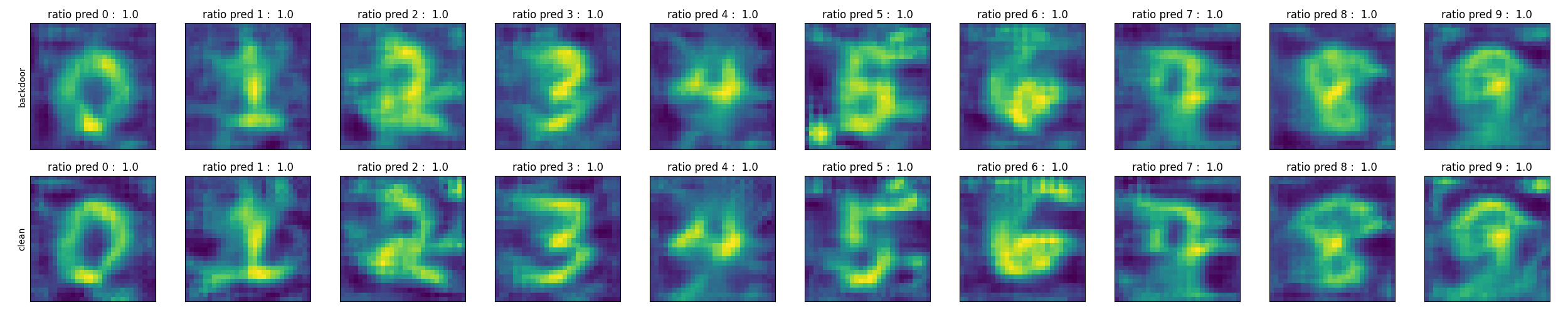}
\includegraphics[width=1\linewidth]{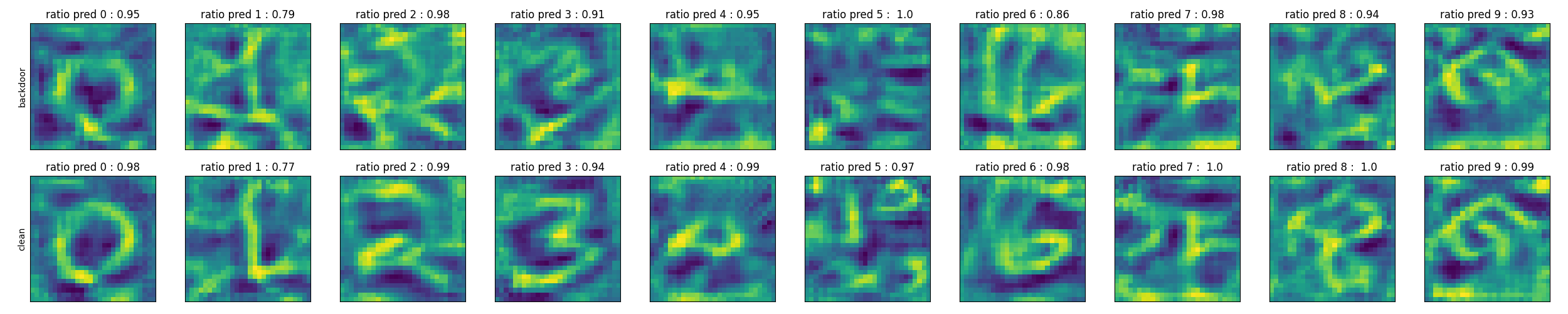}
\vspace{15pt}
\includegraphics[width=1\linewidth]{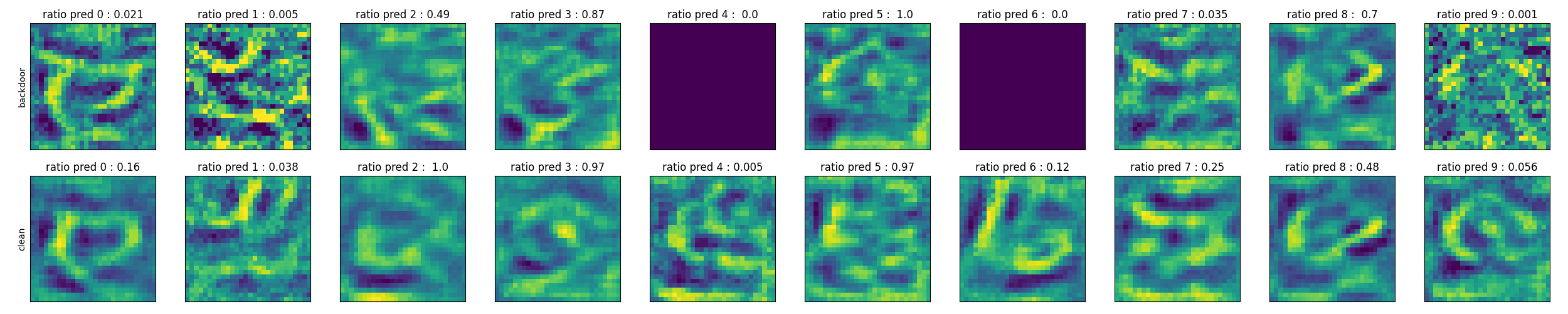}

\ \ \ \ Using noise \\
\includegraphics[width=1\linewidth]{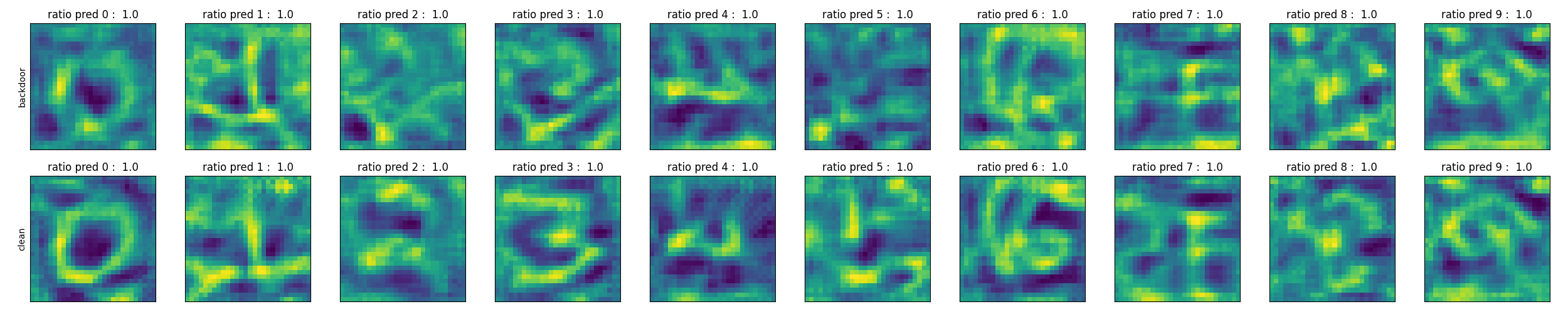}
\includegraphics[width=1\linewidth]{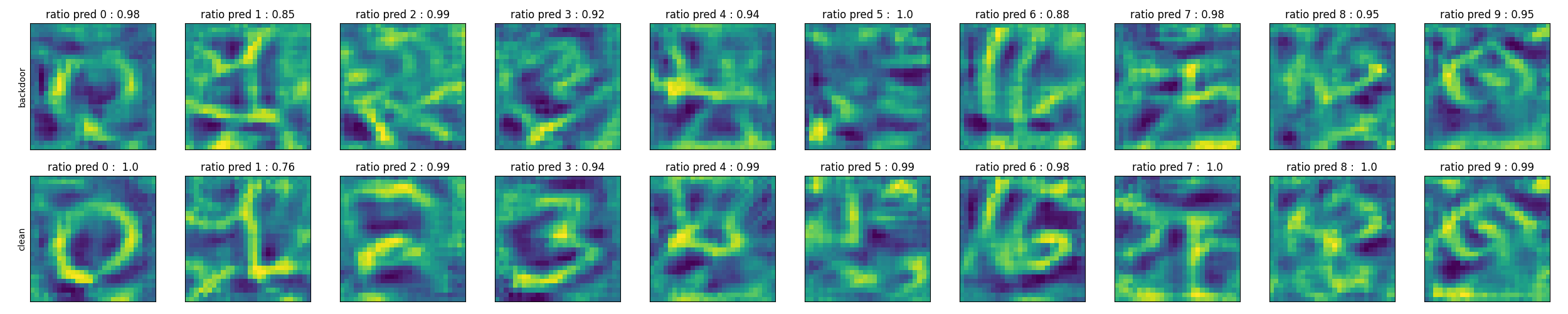}
\vspace{15pt}
\includegraphics[width=1\linewidth]{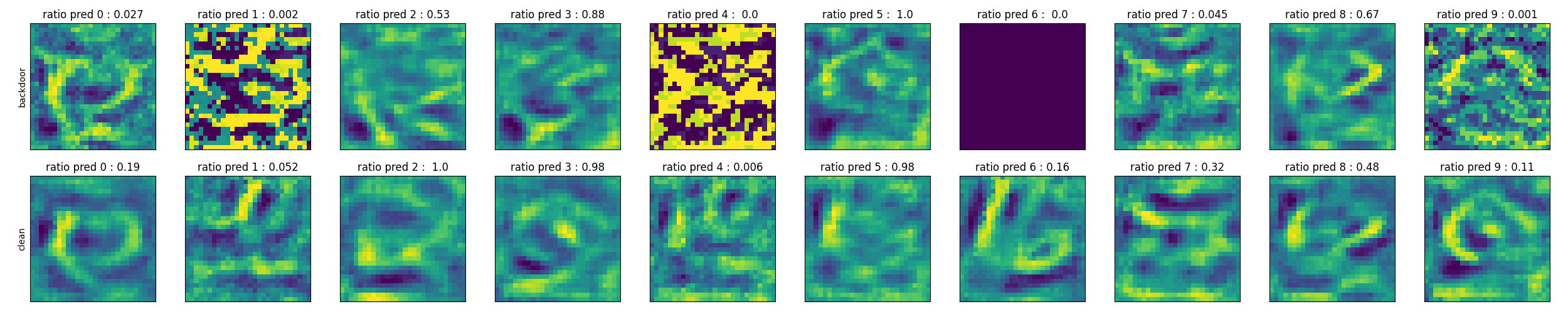}
\vspace{-30pt}
\caption{Bias maps derived by feeding data (top) and white noise (bottom) to Algorithm I, for the $2 \rightarrow 5$ attack. In each panel, pairs from top to bottom correspond to $\epsilon$ equal to 0.1, 0.5, and 0.9, respectively. Number of samples and iterations were set to 1000, 10, respectively.}
\label{fig:25}
\end{adjustwidth}
\end{figure}

Results using data and white noise inputs for the $8\rightarrow9$ are shown in Fig~\ref{fig:89}. Bias maps look more enticing now especially using data as input and with small $\epsilon=0.1$. Again, maps look more or less similar across unaffected categories. Similar results over other attacks are observed as shown in Fig.~\ref{fig:03} for $0\rightarrow3$, Fig.~\ref{fig:47} for $4\rightarrow7$, and Fig.~\ref{fig:25} for $2\rightarrow5$ attacks.   

Visual inspection works but is not feasible when number of categories is high. Is it possible to devise some quantitative measures? To this end, I computed the Euclidean distance between the bias maps of backdoor and clean models. Results are shown in Fig.~\ref{fig:res}. Interestingly, the difference curves peak at either the source or target categories. I expect even better 
results using higher number of adversarial examples. Here, I only generated 1000 examples per category.

Bias maps in Algorithm I only reveal the target category. Lets see if Algorithm II can reveal both source and target categories. Results are shown in Figs~\ref{fig:paired89},~\ref{fig:paired03},~\ref{fig:paired47}, and~\ref{fig:paired25}. Please look at the Figs~\ref{fig:paired89} ($8\rightarrow9$) as an example. Rows and columns in this figure correspond to source and target categories, respectively. Perturbing images, noise or data, not to belong to any source category but 9, highlights the C-shaped adversarial patch (the right most columns). This makes sense since activating this region increases the probability of class 9. With the same token, perturbing images not to be 8 sometimes highlights the patch region. This is perhaps due to the optimization procedure that aims to lower the probability of 8 while at the same time increasing the probability of the target class. Nonetheless, looking across rows and columns reveals which source category has been attacked to become which target category. 

\subsection{Experiment II}
\begin{figure}
\centering
\begin{adjustwidth}{-1cm}{-3cm}
\subfigure{\includegraphics[width=.22\linewidth]{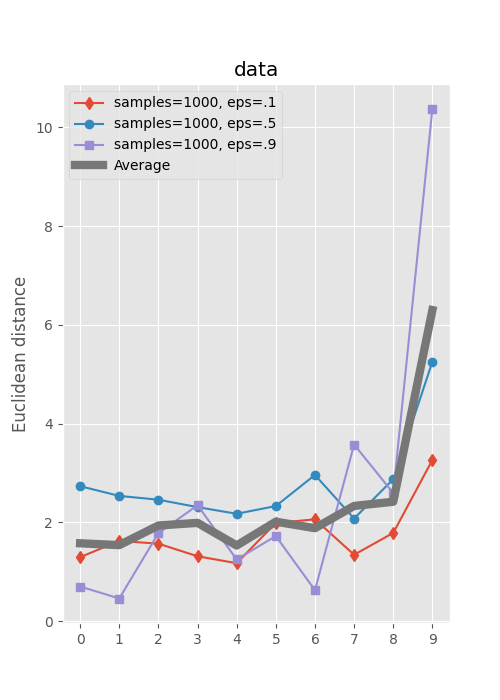}}
\subfigure{\includegraphics[width=.22\linewidth]{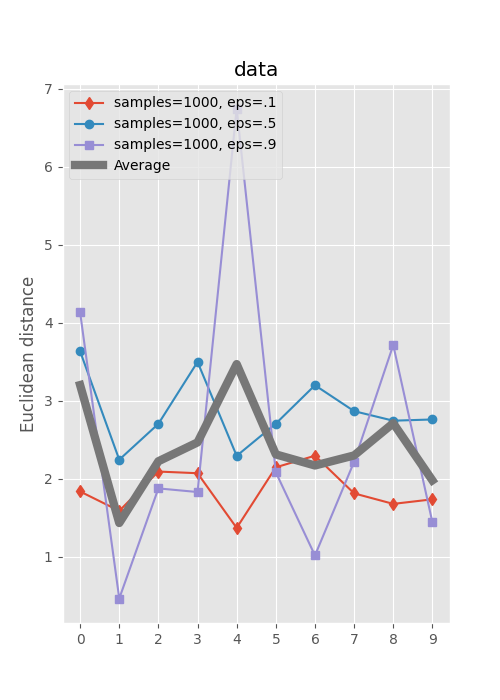}}
\subfigure{\includegraphics[width=.22\linewidth]{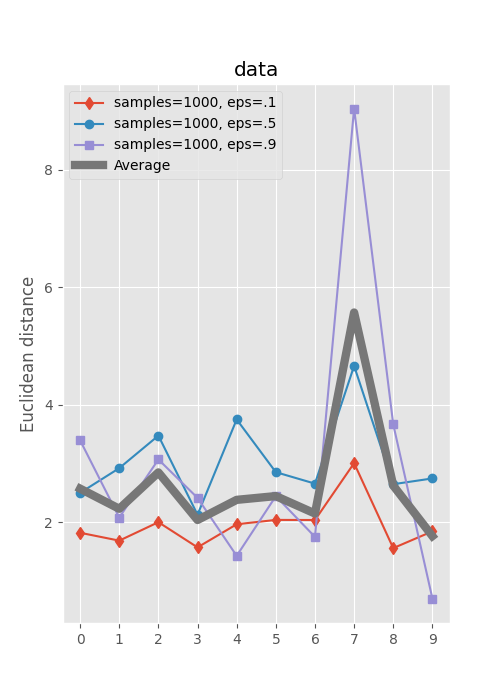}}
\subfigure{\includegraphics[width=.22\linewidth]{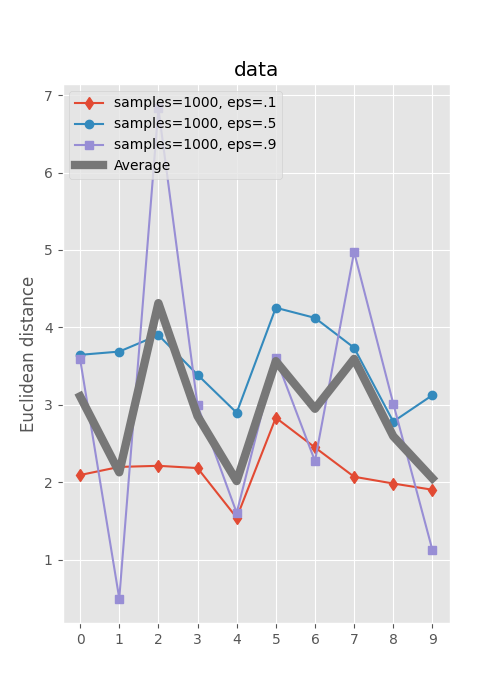}}\\
\subfigure{\includegraphics[width=.22\linewidth]{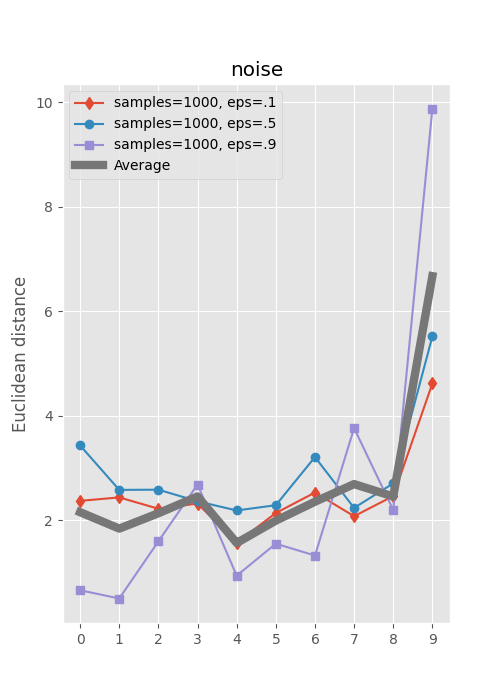}}
\subfigure{\includegraphics[width=.22\linewidth]{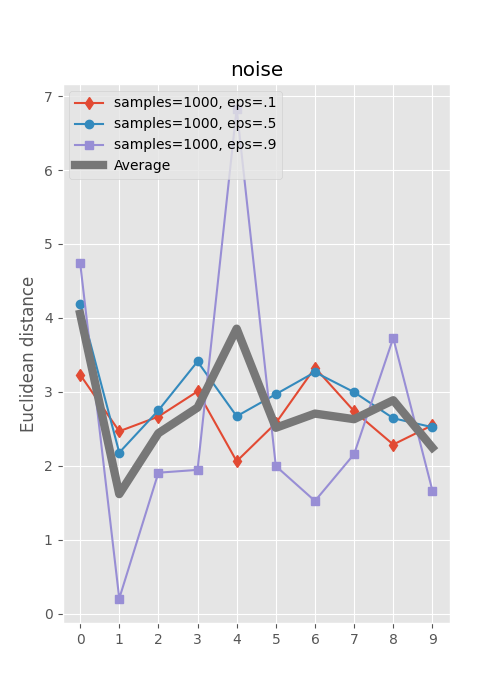}}
\subfigure{\includegraphics[width=.22\linewidth]{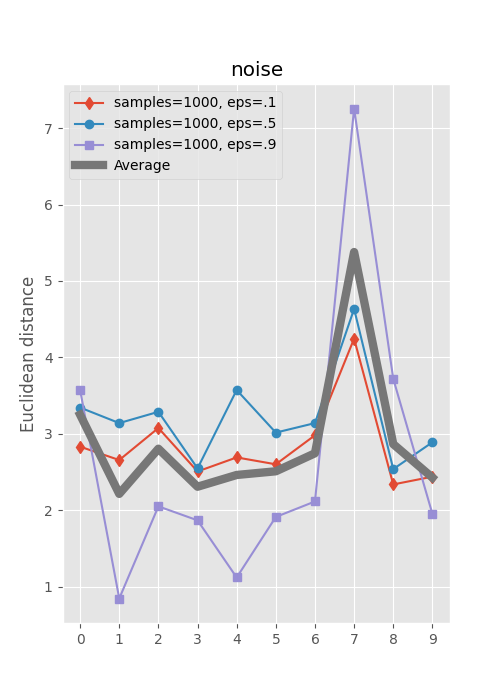}}
\subfigure{\includegraphics[width=.22\linewidth]{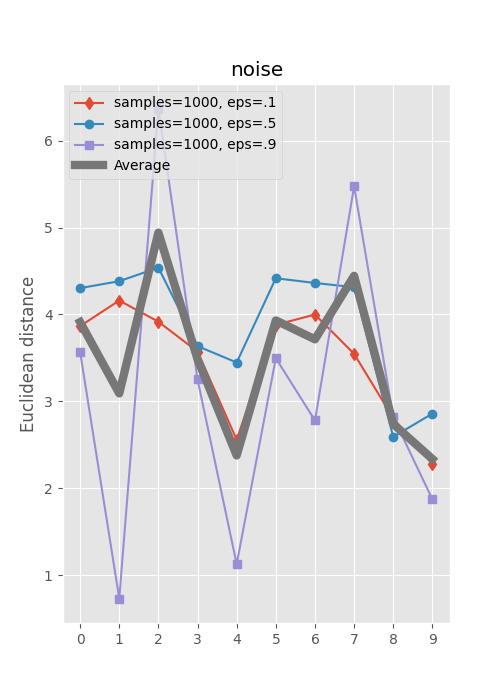}}
\caption{Quantitative performance of Algorithm I in detecting the backdoor attacks in experiment one. \\ Columns from left to right correspond to $8\rightarrow9$, $0\rightarrow3$, $4\rightarrow7$, and $2\rightarrow5$ attacks.}
\label{fig:res}
\end{adjustwidth}    
\end{figure}

\begin{figure}
\centering
\begin{adjustwidth}{-1cm}{-1cm}
\includegraphics[width=.5\linewidth]{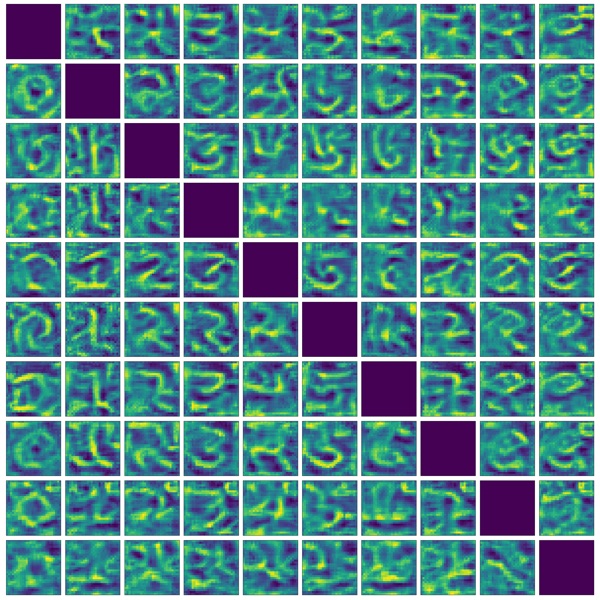}
\includegraphics[width=.5\linewidth]{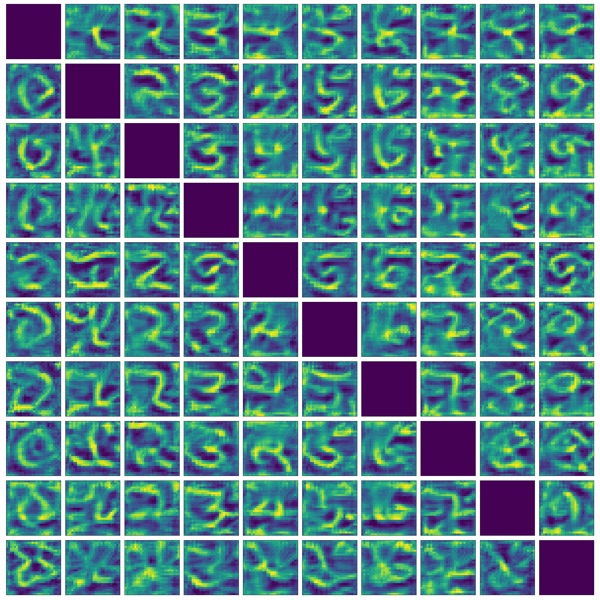}
\\
\includegraphics[width=.5\linewidth]{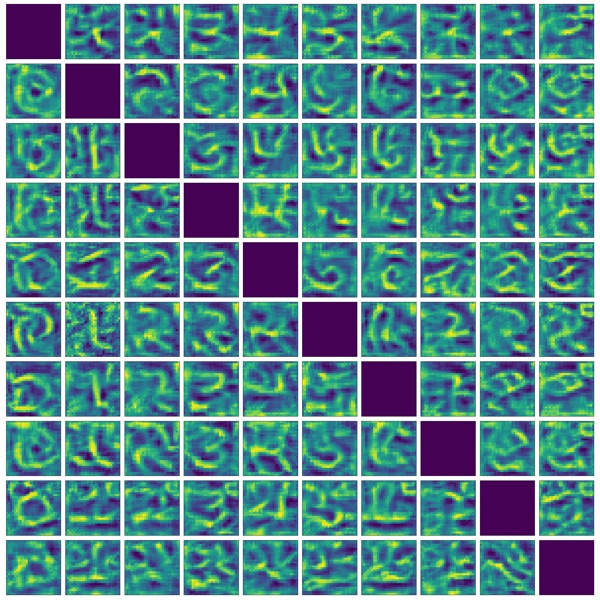}
\includegraphics[width=.5\linewidth]{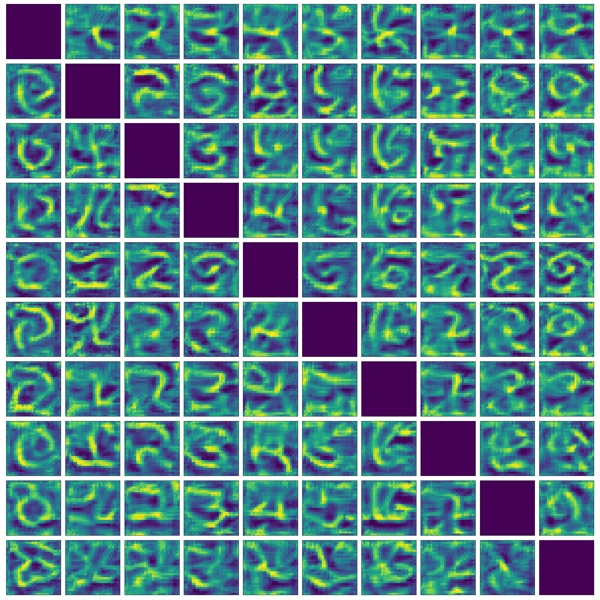}
\caption{Bias maps over pairs of (source,target) categories using data (top row) and white noise (bottom row) inputs. Columns correspond to the backdoor (left) and the clean (right) models. This result is for the $8\rightarrow9$ attack.}
\label{fig:paired89}
\end{adjustwidth}    
\end{figure}

\begin{figure}
\centering
\begin{adjustwidth}{-1cm}{-1cm}
\includegraphics[width=.5\linewidth]{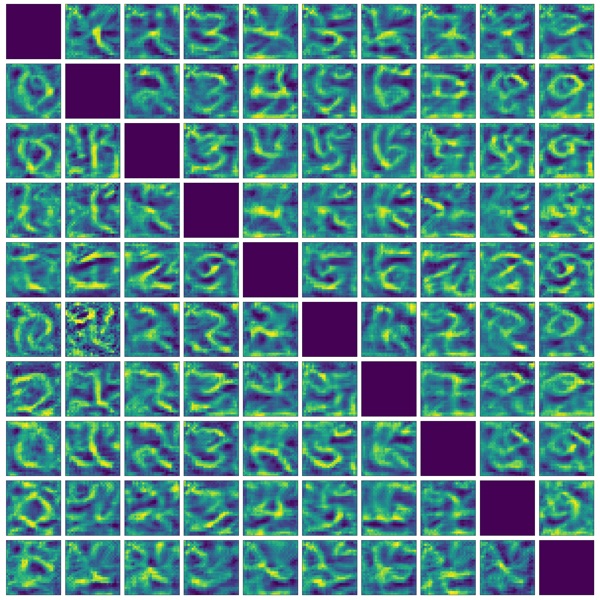}
\includegraphics[width=.5\linewidth]{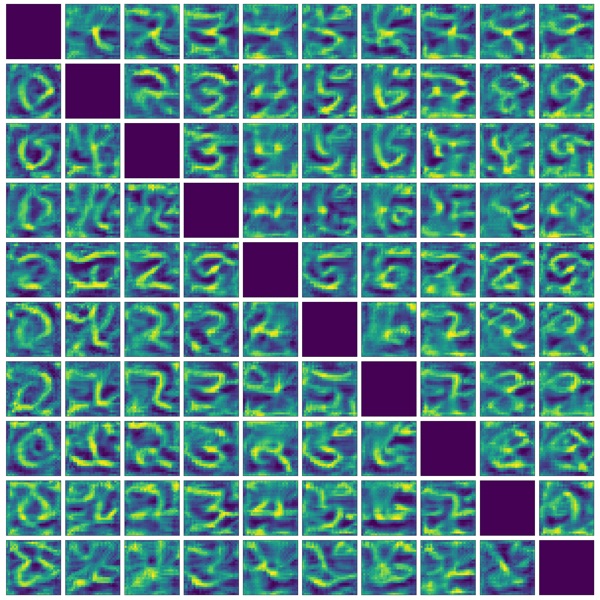}
\\
\includegraphics[width=.5\linewidth]{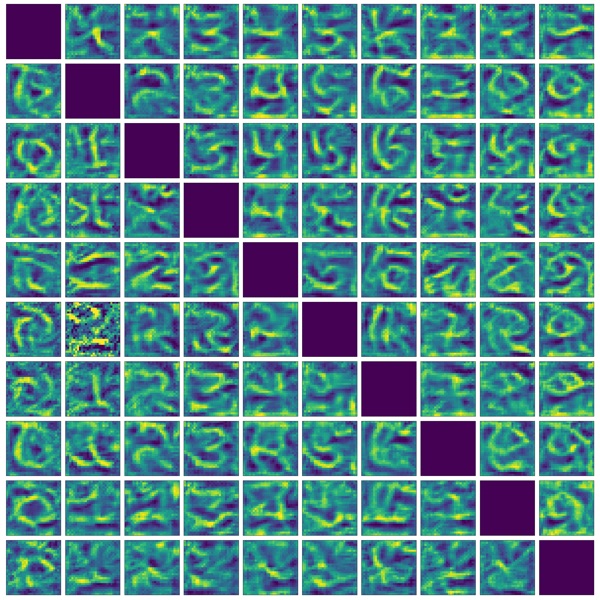}
\includegraphics[width=.5\linewidth]{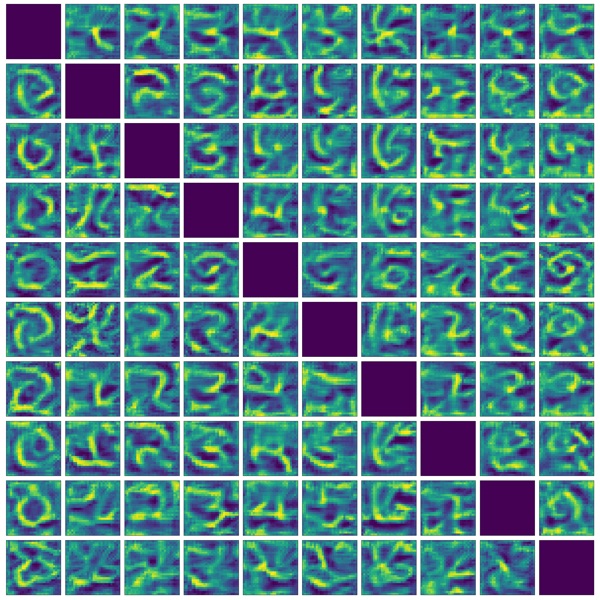}
\caption{Bias maps over pairs of (source,target) categories using data (top row) and white noise (bottom row) inputs. Columns correspond to the backdoor (left) and the clean (right) models. This result is for the $0\rightarrow3$ attack.}
\label{fig:paired03}
\end{adjustwidth}    
\end{figure}

\begin{figure}
\centering
\begin{adjustwidth}{-1cm}{-1cm}
\includegraphics[width=.5\linewidth]{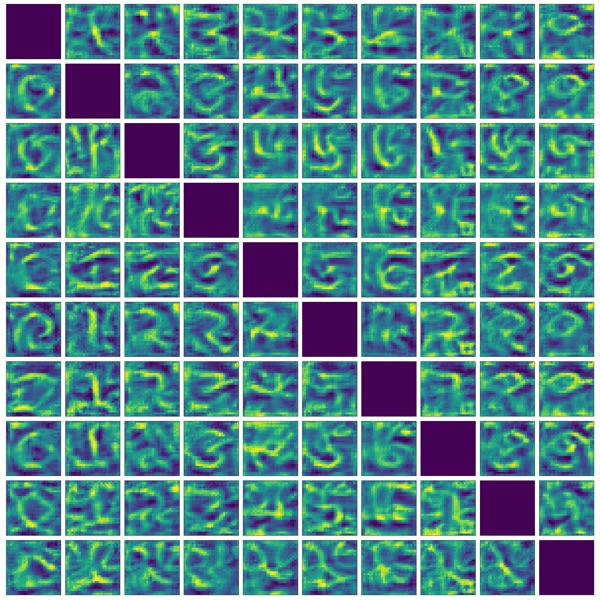}
\includegraphics[width=.5\linewidth]{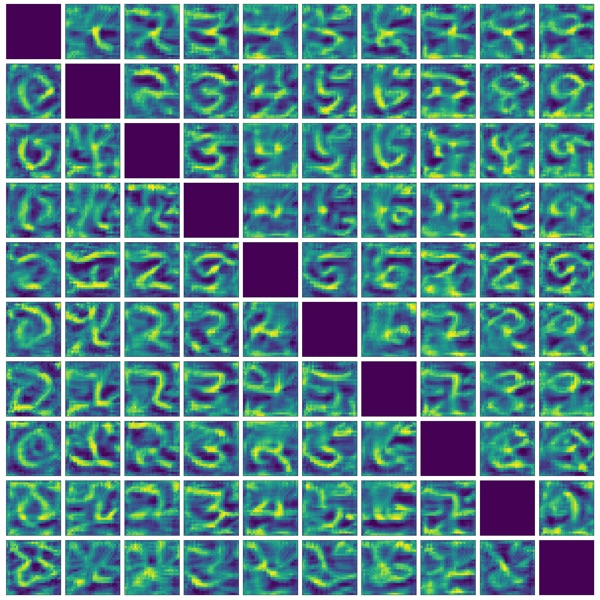}
\\
\includegraphics[width=.5\linewidth]{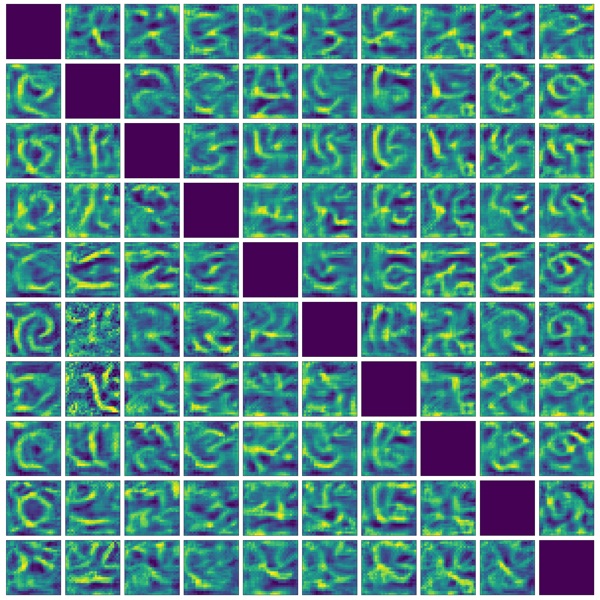}
\includegraphics[width=.5\linewidth]{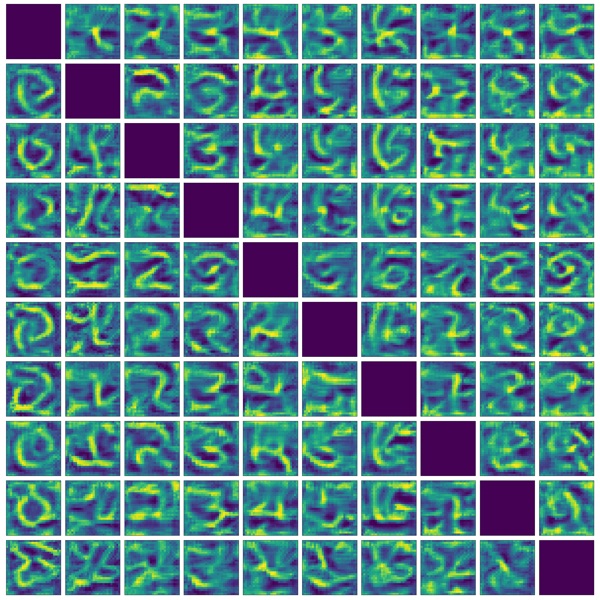}
\caption{Bias maps over pairs of (source,target) categories using data (top row) and white noise (bottom row) inputs. Columns correspond to the backdoor (left) and the clean (right) models. This result is for the $4\rightarrow7$ attack.}
\label{fig:paired47}
\end{adjustwidth}    
\end{figure}

\begin{figure}
\centering
\begin{adjustwidth}{-1cm}{-1cm}
\includegraphics[width=.5\linewidth]{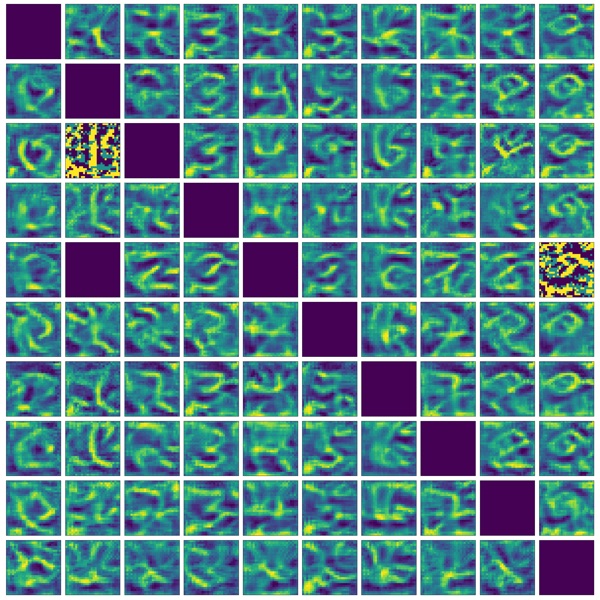}
\includegraphics[width=.5\linewidth]{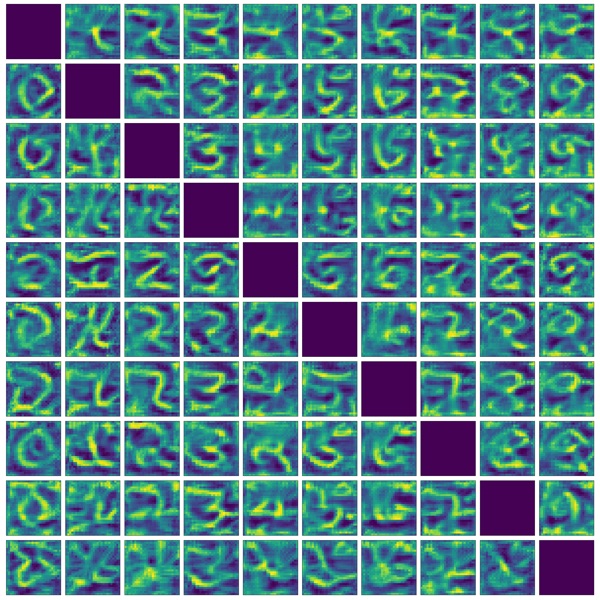}
\\
\includegraphics[width=.5\linewidth]{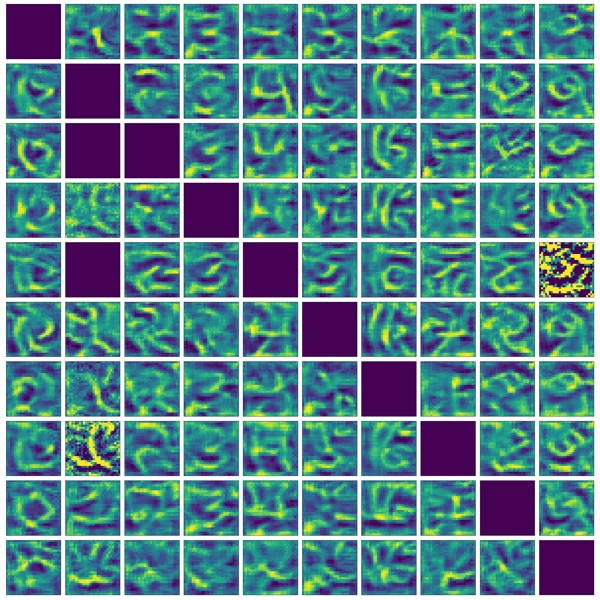}
\includegraphics[width=.5\linewidth]{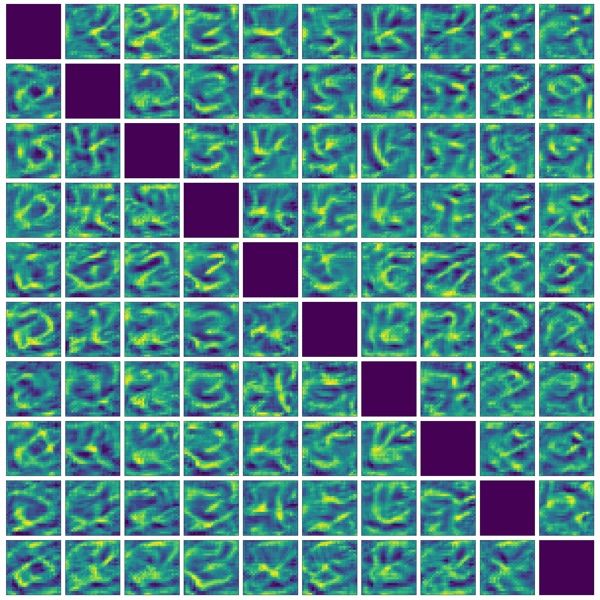}
\caption{Bias maps over pairs of (source,target) categories using data (top row) and white noise (bottom row) inputs. Columns correspond to the backdoor (left) and the clean (right) models. This result is for the $2\rightarrow5$ attack.}
\label{fig:paired25}
\end{adjustwidth}    
\end{figure}

\begin{figure}[htbp]       
\begin{adjustwidth}{-2cm}{-2cm}
\ \ \ \ Using data \\
\includegraphics[width=1\linewidth]{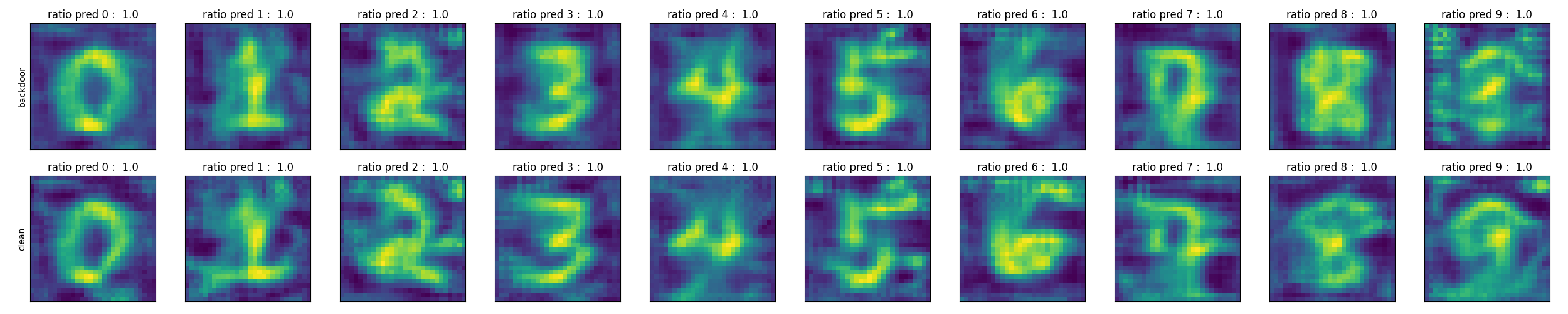}
\vspace{10pt}
\includegraphics[width=1\linewidth]{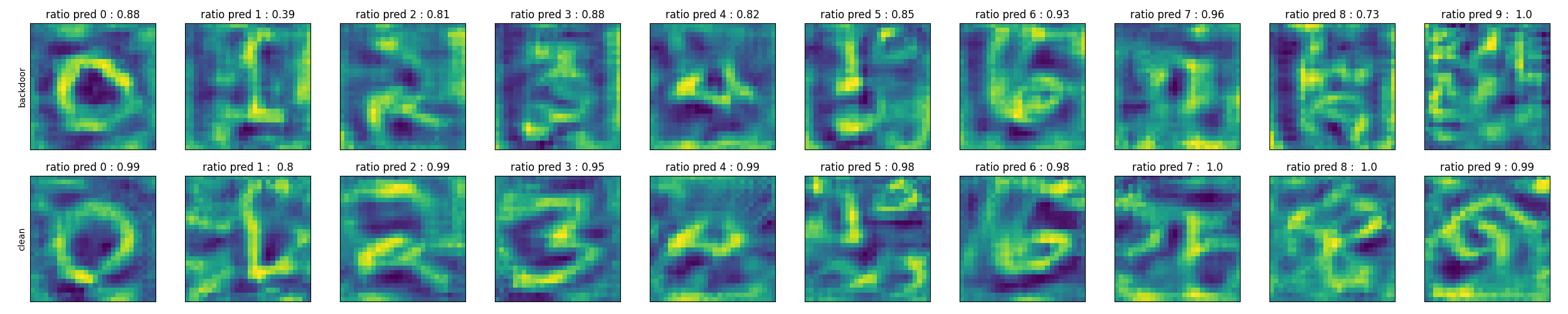}

\ \ \ \ Using noise \\
\includegraphics[width=1\linewidth]{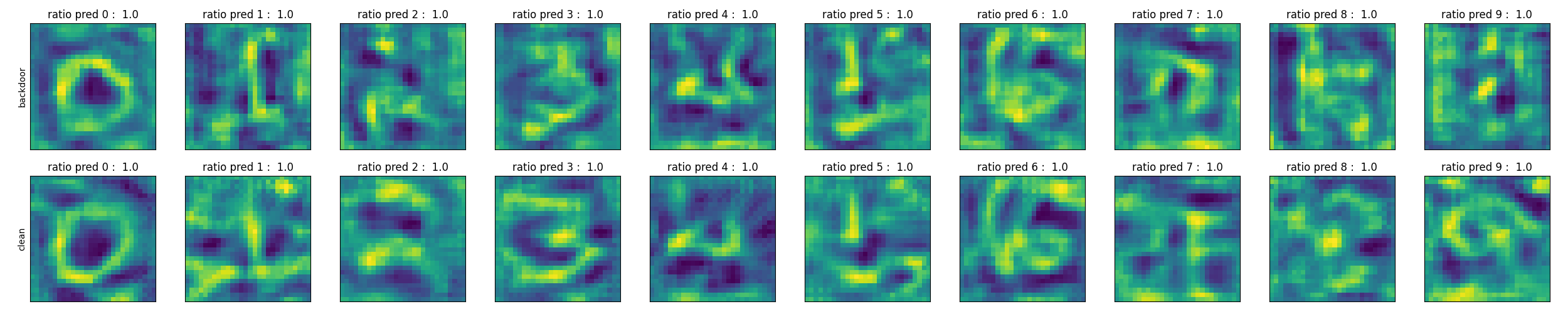}
\vspace{10pt}
\includegraphics[width=1\linewidth]{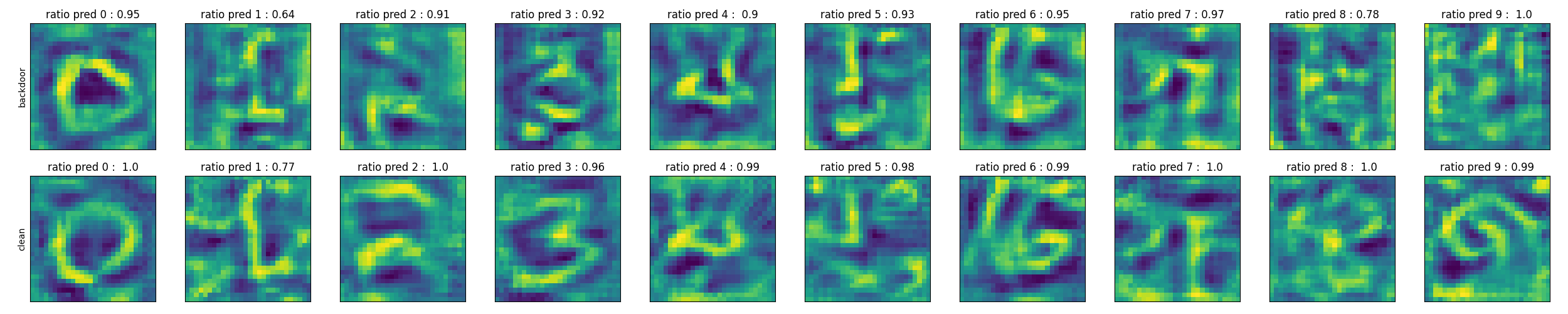}

\caption{Bias maps derived by feeding data (top) and white noise (bottom) to Algorithm I, for the $8 \rightarrow 9$ attack. In each panel, pairs from top to bottom correspond to $\epsilon$ equal to 0.1 and 0.5 respectively. Number of samples and iterations were set to 1000 and 10, respectively. Adversarial patches were placed at random locations.}
\label{fig:89loc}
\end{adjustwidth}
\end{figure}

\begin{figure}[htbp]       
\begin{adjustwidth}{-2cm}{-2cm}
\ \ \ \ Using data \\
\includegraphics[width=1\linewidth]{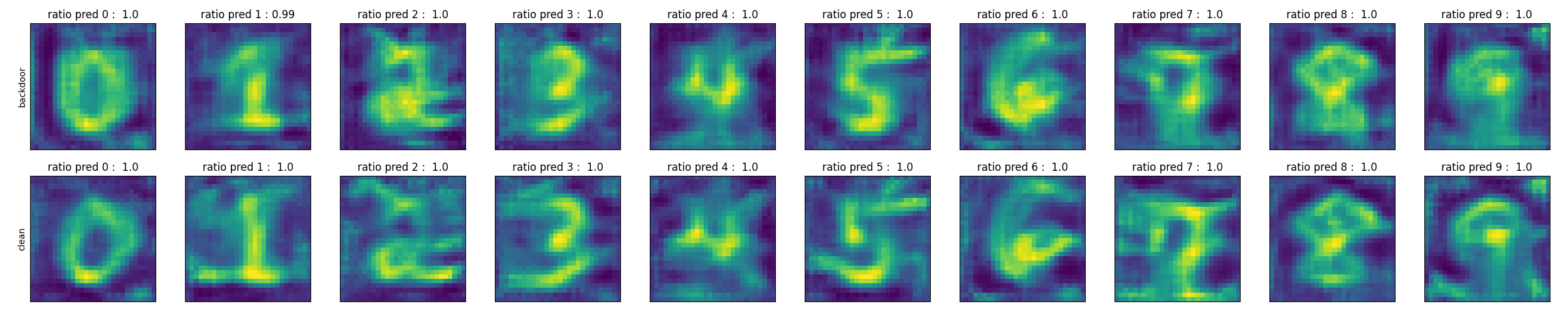}
\vspace{10pt}
\includegraphics[width=1\linewidth]{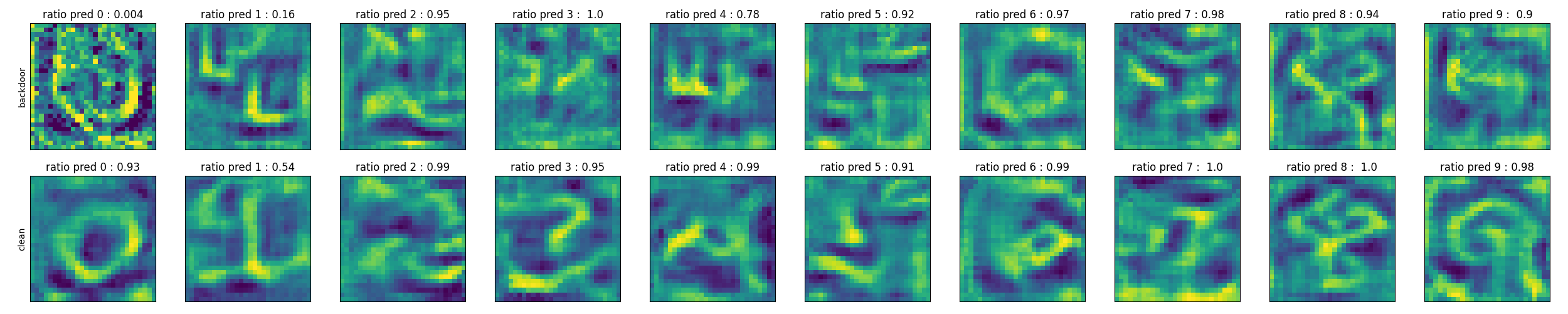}

\ \ \ \ Using noise \\
\includegraphics[width=1\linewidth]{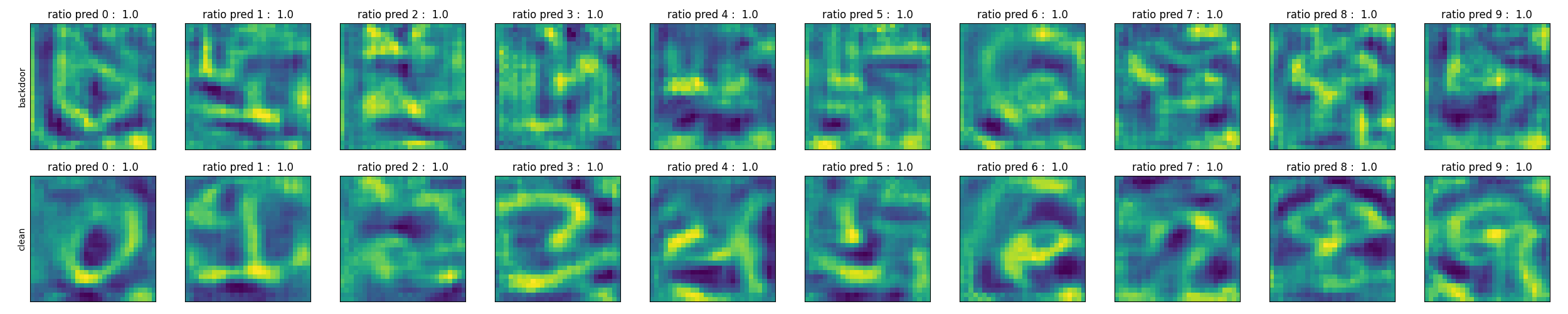}
\vspace{10pt}
\includegraphics[width=1\linewidth]{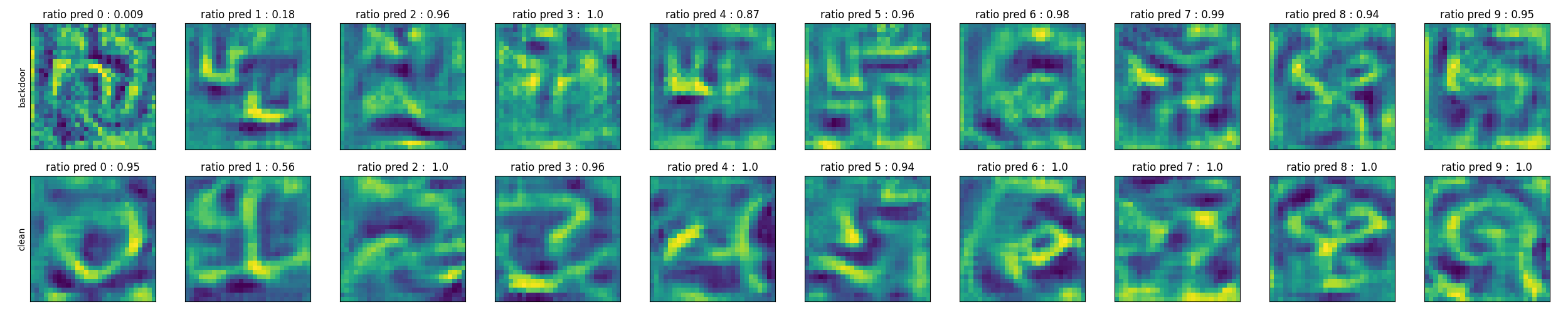}

\caption{Bias maps derived by feeding data (top) and white noise (bottom) to Algorithm I, for the $0 \rightarrow 3$ attack. In each panel, pairs from top to bottom correspond to $\epsilon$ equal to 0.1 and 0.5 respectively. Number of samples and iterations were set to 1000 and 10, respectively. Adversarial patches were placed at random locations.}
\label{fig:03loc}
\end{adjustwidth}
\end{figure}

\begin{figure}[htbp]       
\begin{adjustwidth}{-2cm}{-2cm}
\ \ \ \ Using data \\
\includegraphics[width=1\linewidth]{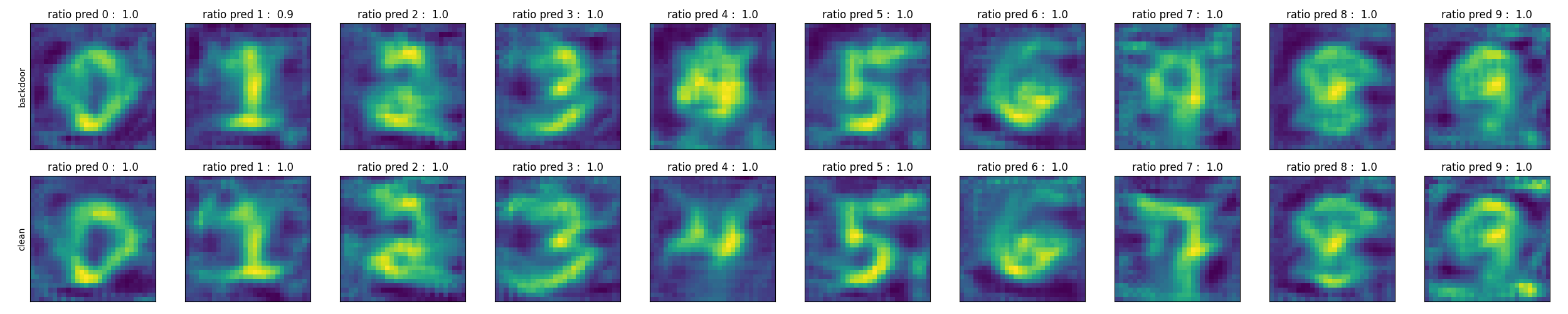}
\vspace{10pt}
\includegraphics[width=1\linewidth]{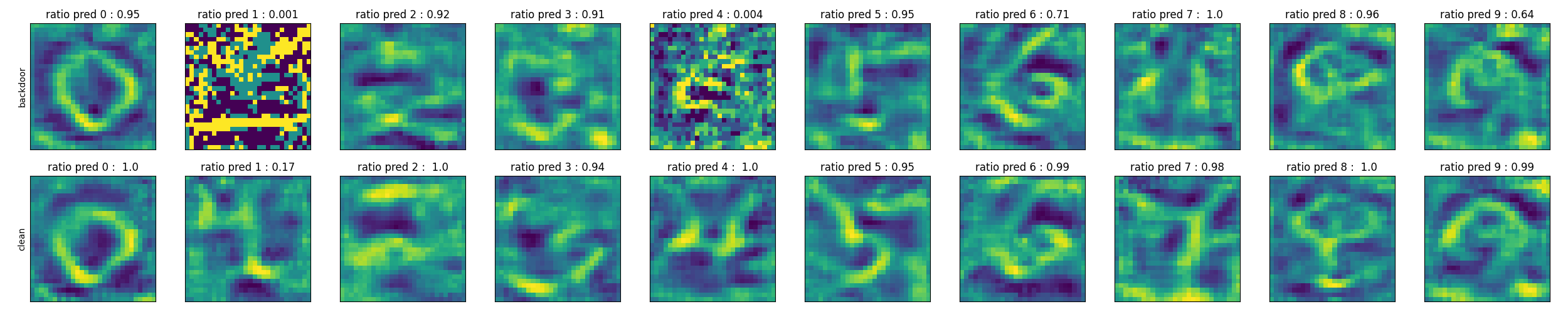}

\ \ \ \ Using noise \\
\includegraphics[width=1\linewidth]{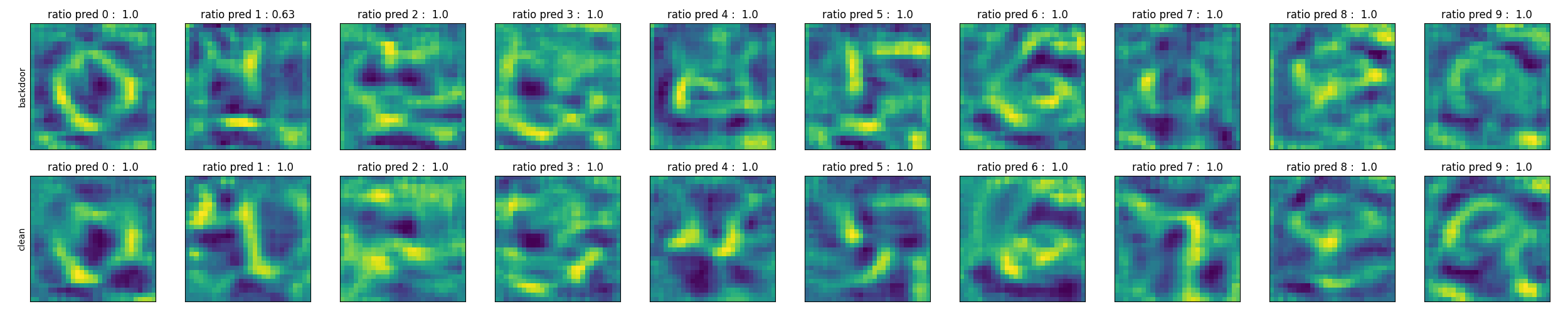}
\vspace{10pt}
\includegraphics[width=1\linewidth]{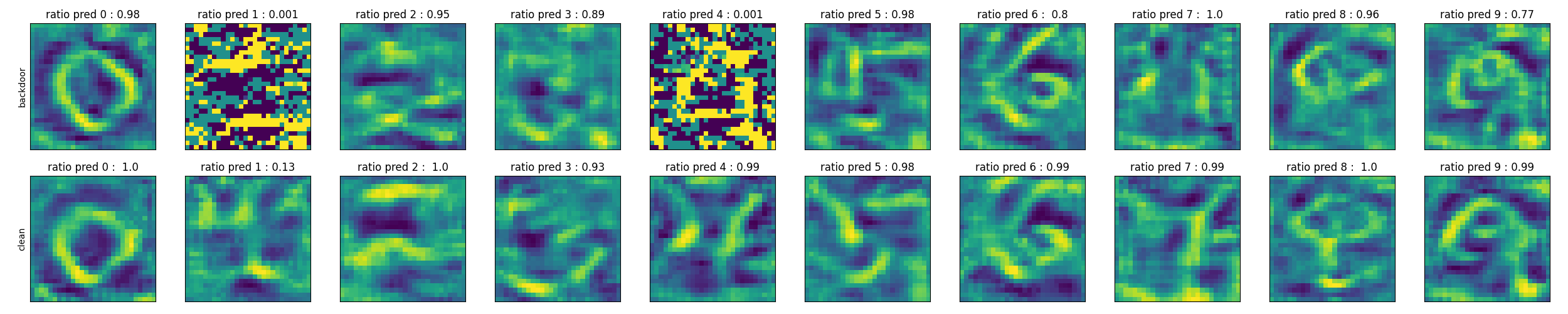}

\caption{Bias maps derived by feeding data (top) and white noise (bottom) to Algorithm I, for the $4 \rightarrow 7$ attack. In each panel, pairs from top to bottom correspond to $\epsilon$ equal to 0.1 and 0.5 respectively. Number of samples and iterations were set to 1000 and 10, respectively. Adversarial patches were placed at random locations.}
\label{fig:47loc}
\end{adjustwidth}
\end{figure}

\begin{figure}[htbp]       
\begin{adjustwidth}{-2cm}{-2cm}
\ \ \ \ Using data \\
\includegraphics[width=1\linewidth]{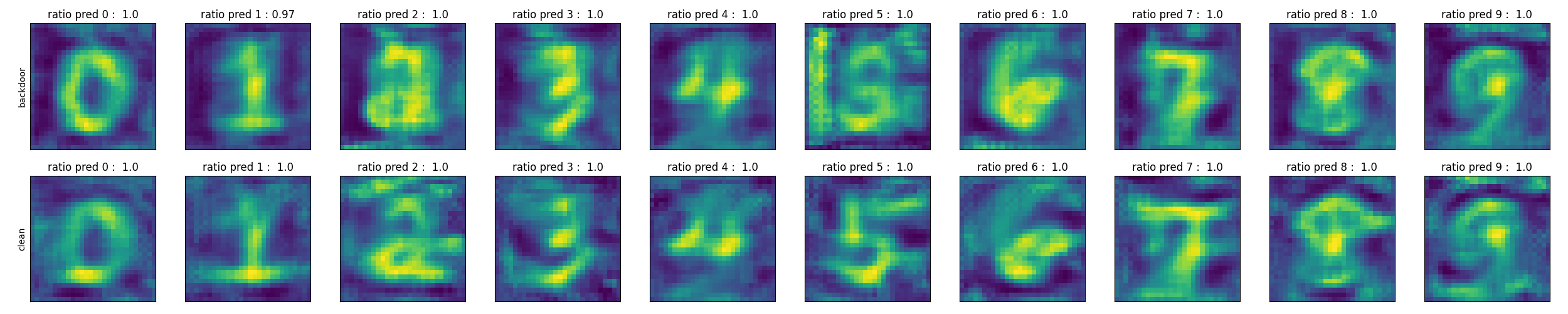}
\vspace{10pt}
\includegraphics[width=1\linewidth]{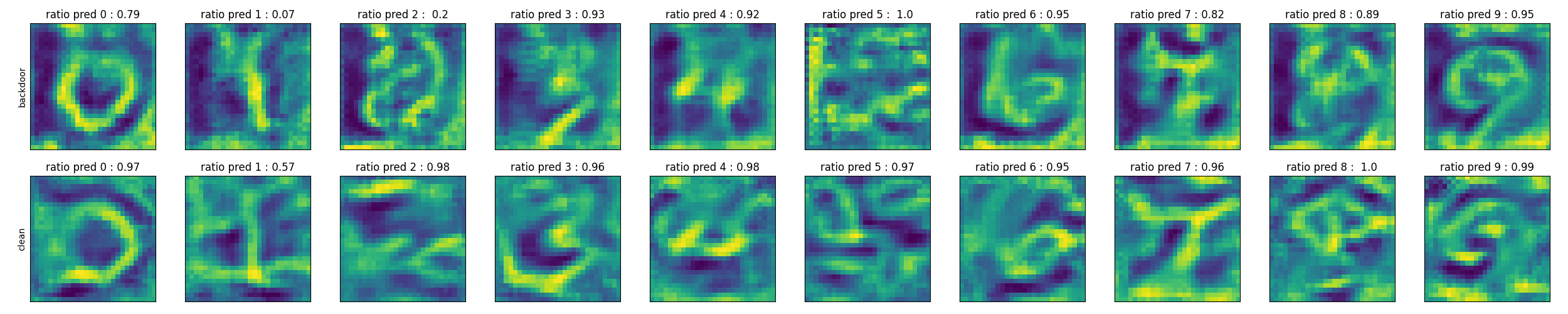}

\ \ \ \ Using noise \\
\includegraphics[width=1\linewidth]{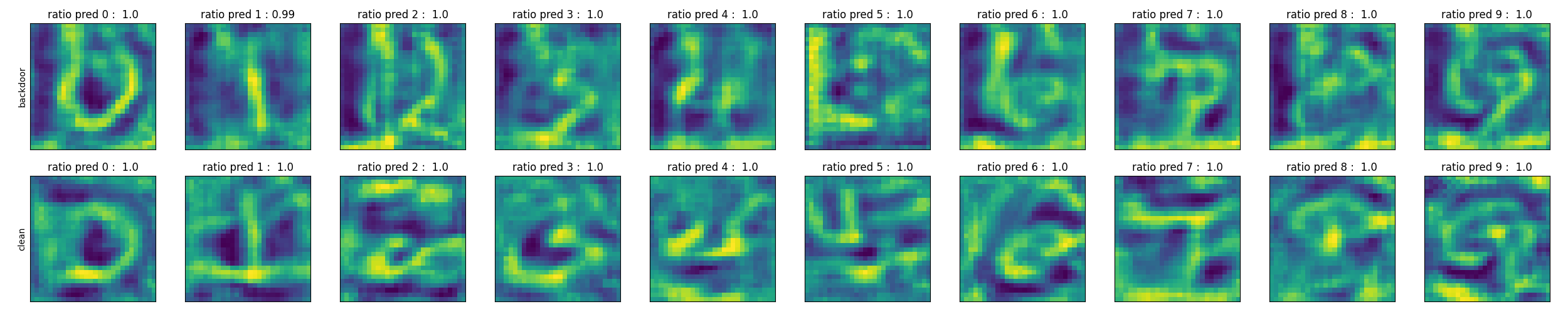}
\vspace{10pt}
\includegraphics[width=1\linewidth]{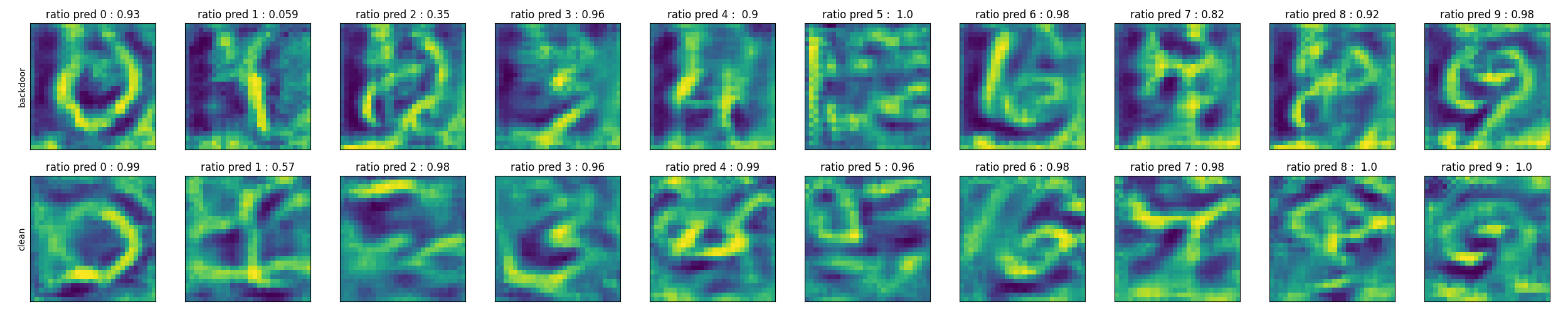}

\caption{Bias maps derived by feeding data (top) and white noise (bottom) to Algorithm I, for the $2 \rightarrow 5$ attack. In each panel, pairs from top to bottom correspond to $\epsilon$ equal to 0.1 and 0.5 respectively. Number of samples and iterations were set to 1000 and 10, respectively. Adversarial patches were placed at random locations.}
\label{fig:25loc}
\end{adjustwidth}
\end{figure}

\begin{figure}
\centering
\begin{adjustwidth}{-1cm}{-3cm}
\subfigure{\includegraphics[width=.22\linewidth]{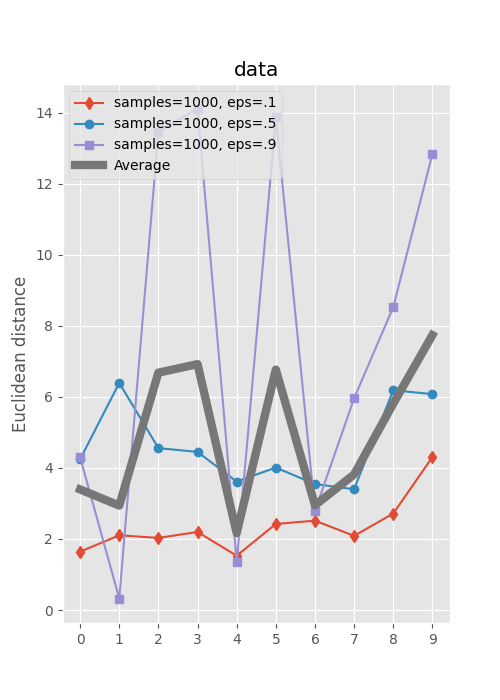}}
\subfigure{\includegraphics[width=.22\linewidth]{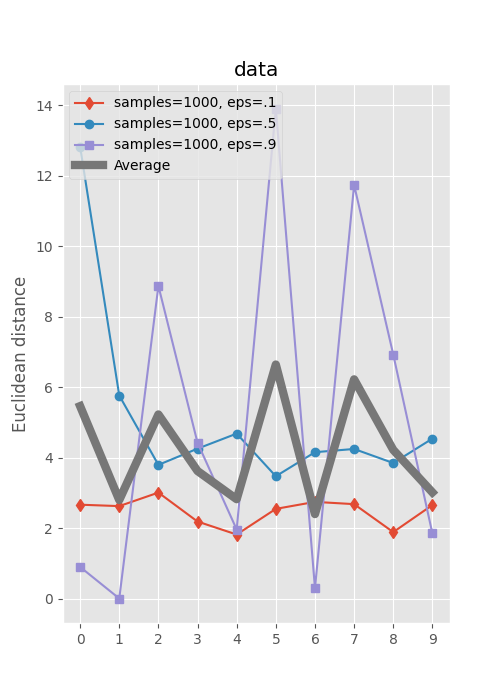}}
\subfigure{\includegraphics[width=.22\linewidth]{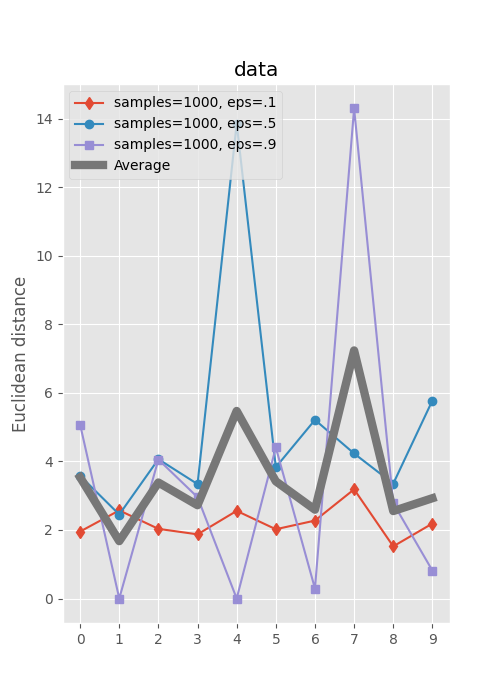}}
\subfigure{\includegraphics[width=.22\linewidth]{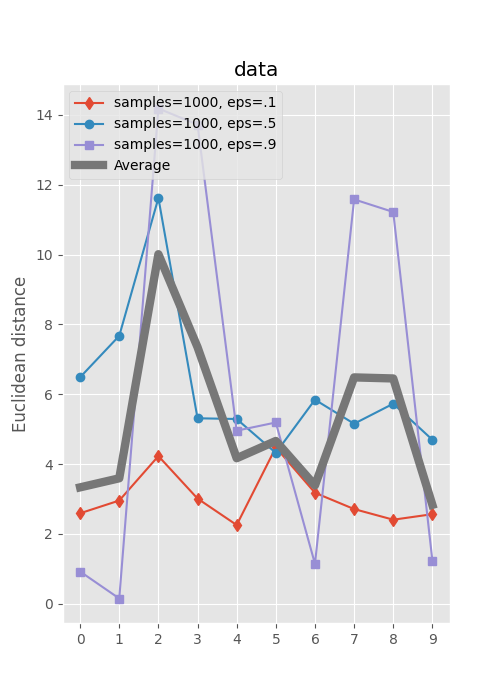}}\\
\subfigure{\includegraphics[width=.22\linewidth]{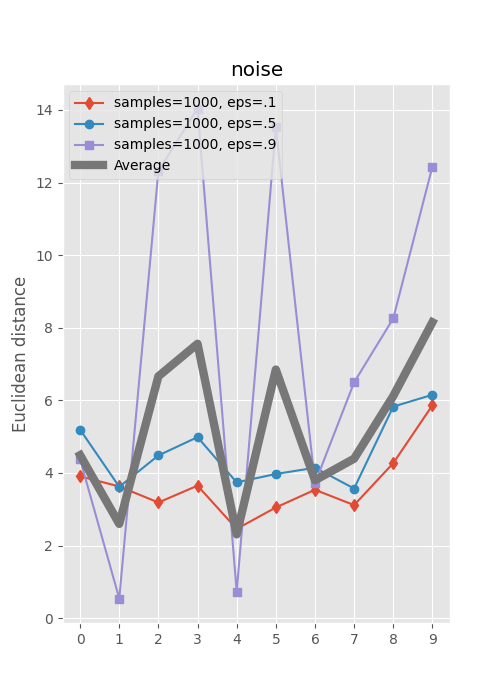}}
\subfigure{\includegraphics[width=.22\linewidth]{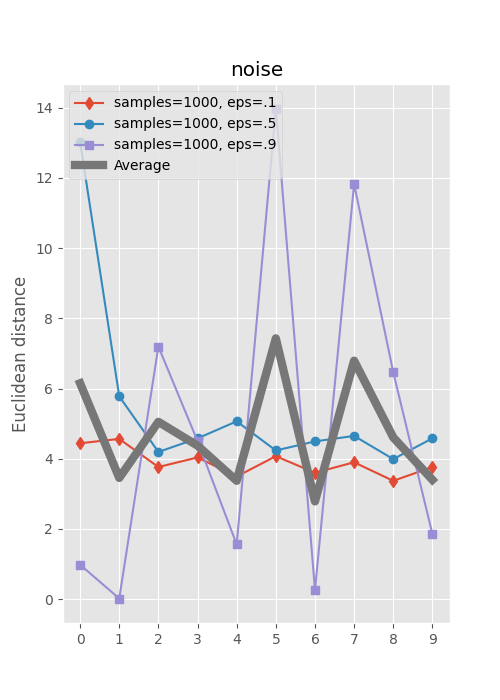}}
\subfigure{\includegraphics[width=.22\linewidth]{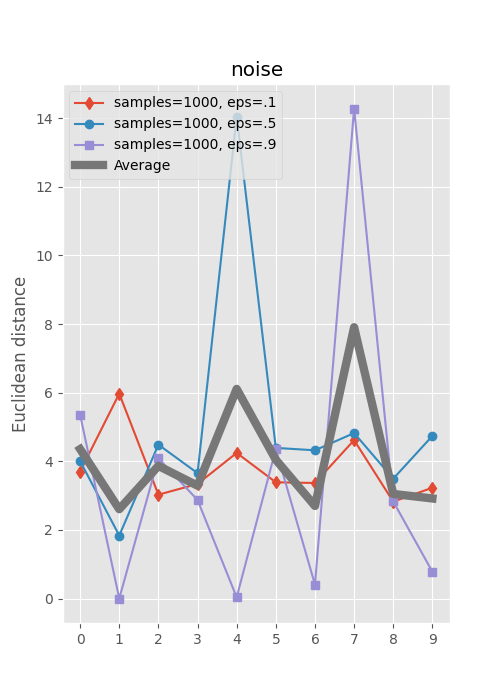}}
\subfigure{\includegraphics[width=.22\linewidth]{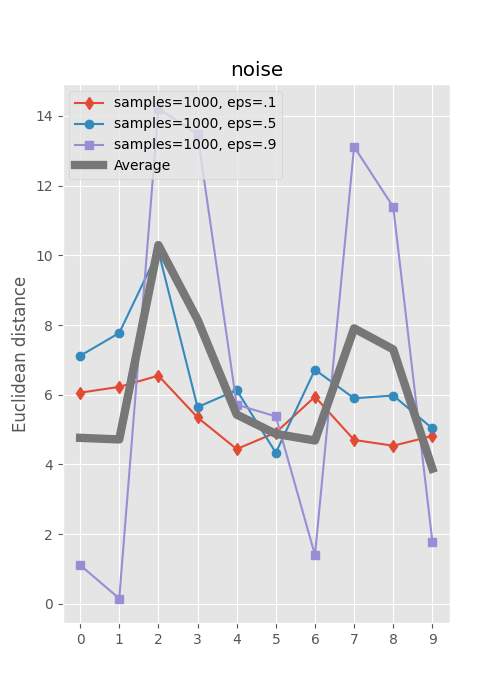}}
\caption{Quantitative performance of Algorithm I in detecting the backdoor attacks in experiment two. Columns from left to right correspond to \\ $8\rightarrow9$, $0\rightarrow3$, $4\rightarrow7$, and $2\rightarrow5$ attacks. Adversarial patches were placed at random locations.}

\label{fig:res-loc}
\end{adjustwidth}    
\end{figure}

Adversarial patches in experiment one were confined to a fixed spatial region. Here, I will choose random locations to place the patch. Attempts were made not to occlude the digits (\ie blank regions were selected). Results are shown in Fig.~\ref{fig:89loc} for the $8\rightarrow9$ attack. Inspecting the bias map for digit 9 shows C-shaped patterns all over the map. Again, results will likely improve with more number of samples per category, since number of possible locations for the patch are now higher compared to the first experiment.
Quantitative results are shown in Fig.~\ref{fig:res-loc}. Bar charts are spikier now compared to Fig.~\ref{fig:res} but still in some cases reveal the impacted categories. Please see Figs.~\ref{fig:03loc},~\ref{fig:47loc}, and~\ref{fig:25loc} for results over other attacks. 

\subsection{Experiment III}
\begin{figure}[htbp]       
    \centering
    \includegraphics[width=.22\linewidth]{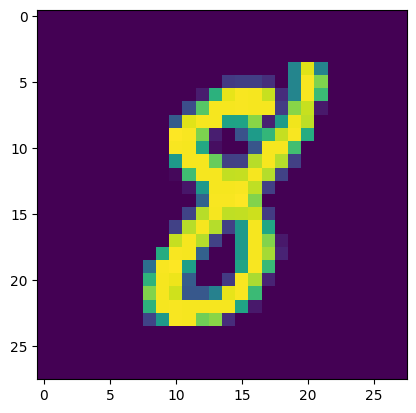} 
    \includegraphics[width=.22\linewidth]{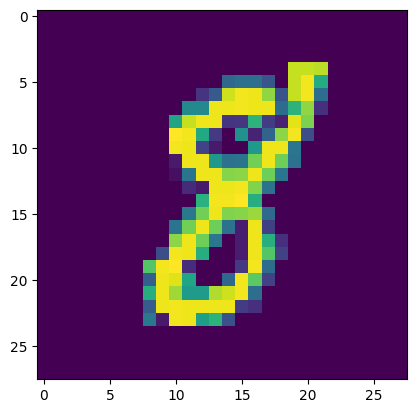}
    \includegraphics[width=.22\linewidth]{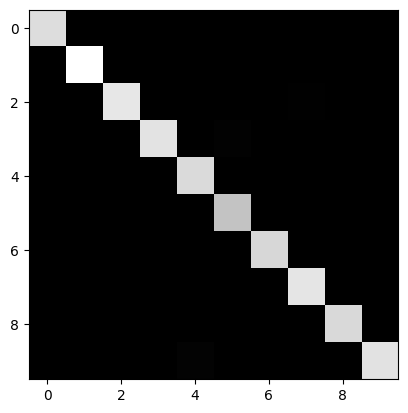}
    \includegraphics[width=.22\linewidth]{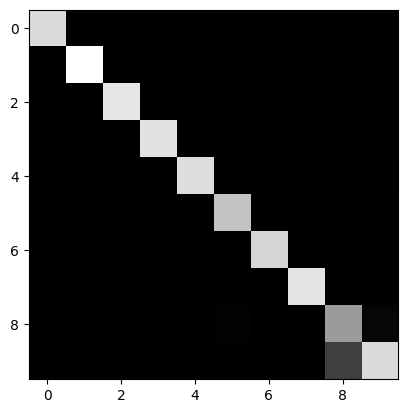} \\
    \vspace{10pt}
    \includegraphics[width=.9\linewidth]{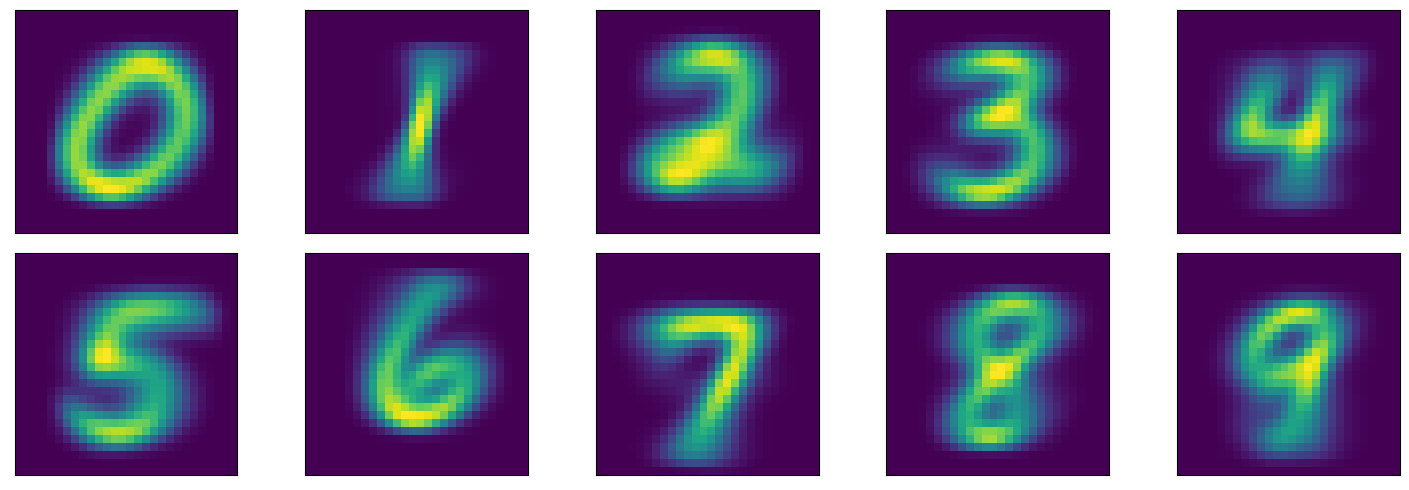}
    \caption{Illustration of the multiplication backdoor attack. A sample image and its corresponding poisoned image, confusion matrices over the pristine testing dataset and the poisoned dataset, as well as the average digit maps are shown.}
\label{fig:minuteMult}
\end{figure}

\begin{figure}[htbp]       
    \centering
    \includegraphics[width=.22\linewidth]{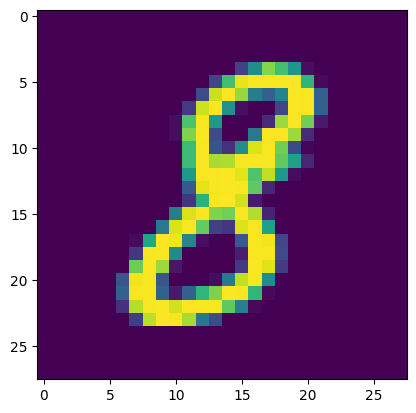} 
    \includegraphics[width=.22\linewidth]{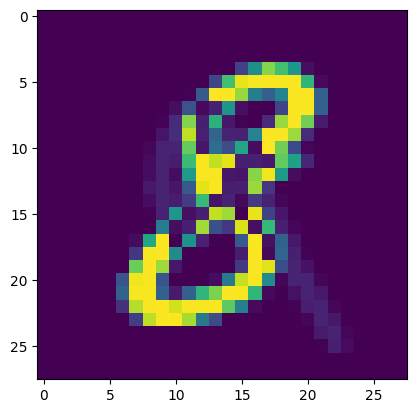}
    \includegraphics[width=.22\linewidth]{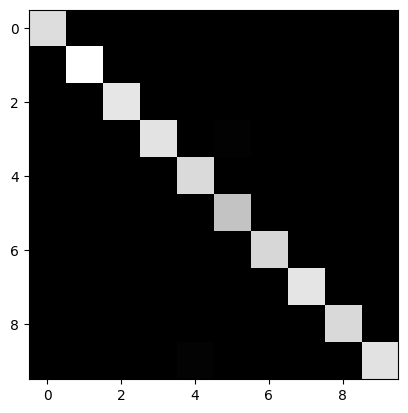}
    \includegraphics[width=.22\linewidth]{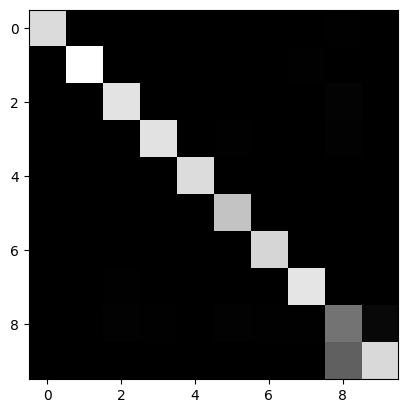} \\
    \vspace{10pt}
    \includegraphics[width=.9\linewidth]{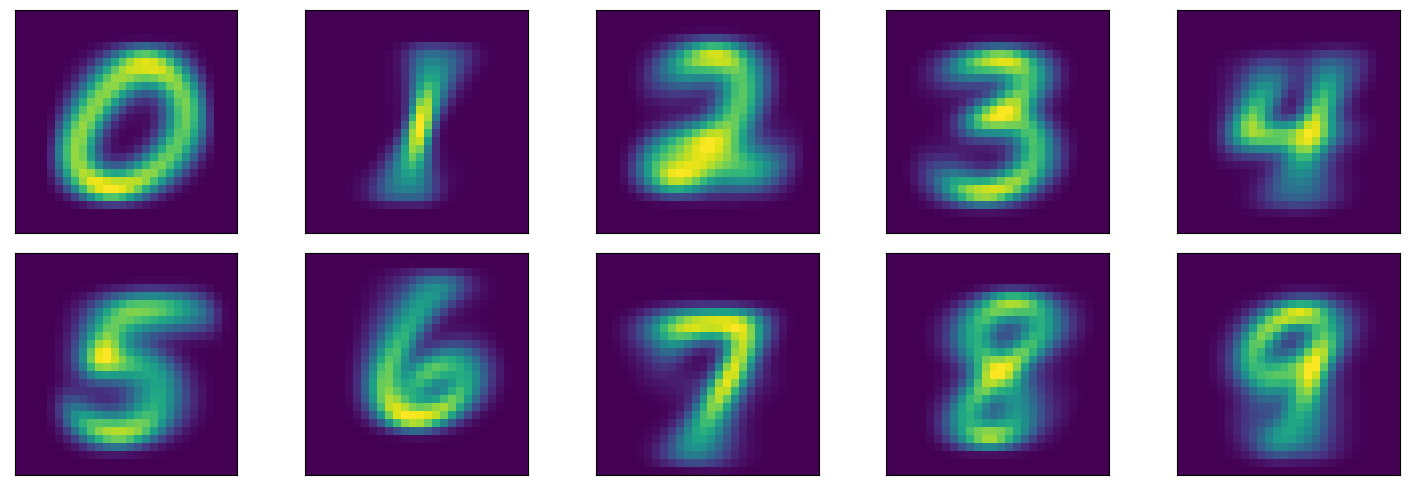}
    \caption{Illustration of the blending backdoor attack. A sample image and its corresponding poisoned image, confusion matrices over the pristine testing dataset and the poisoned dataset, as well as the average digit maps are shown.}
\label{fig:minuteBlend}
\end{figure}

In general, detecting backdoor attacks can be a very daunting task. The adversary may not need to obtain 100\% accuracy for the attack which entails that he can try until he gets a hit (unless there is a limit on the number of attempts). Further, the adversary may add minute perturbations to balance between attack accuracy and the chance of the attack being compromised. My aim in this experiment is to study whether the proposed methods can highlight less perceptible adversarial patches such as multiplying the digit image by 2 (Fig.~\ref{fig:minuteMult}) or blending the source image with a randomly selected image from the target category, here $I_{source} + 0.1\times I_{target}$ (Fig.~\ref{fig:minuteBlend})\footnote{Notice that in general, it is also possible to transform the entire images by transformations such as small rotation, blurring, adding noise, or adding physical objects such as adversarial glasses.}.

Sample images and perturbations are presented for the $8\rightarrow9$ attack in Figs.~\ref{fig:minuteMult} and~\ref{fig:minuteBlend}. As confusion matrices in these figures illustrate, only a tiny fraction of the poisoned images are now classified under the target category. Even the average digit maps here look quite normal. Thus, this type of perturbation can be a serious challenge for any backdoor defense method. 

\begin{figure}[htbp]    
\vspace{-30pt}
\begin{adjustwidth}{-2cm}{-2cm}
\ \ \ \ Using blank image \\
\includegraphics[width=1\linewidth]{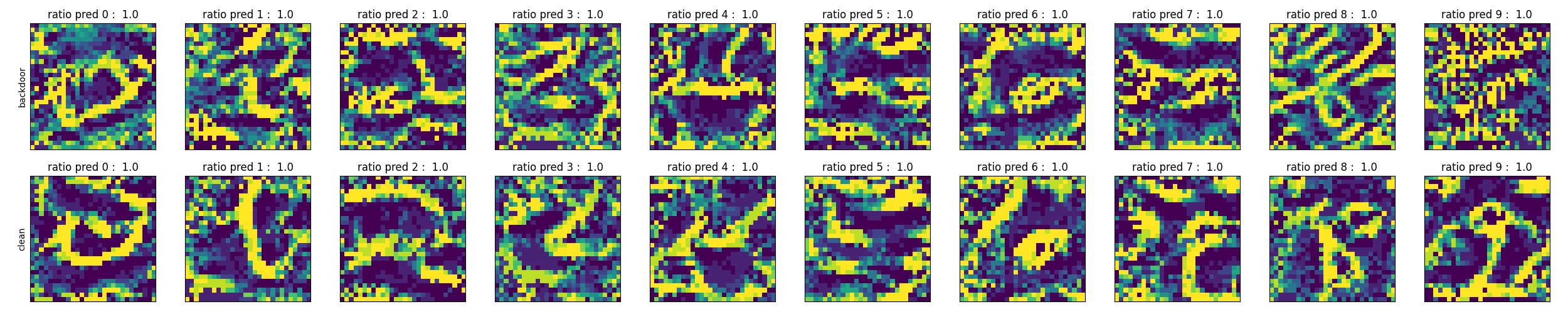}
\vspace{10pt}
\includegraphics[width=1\linewidth]{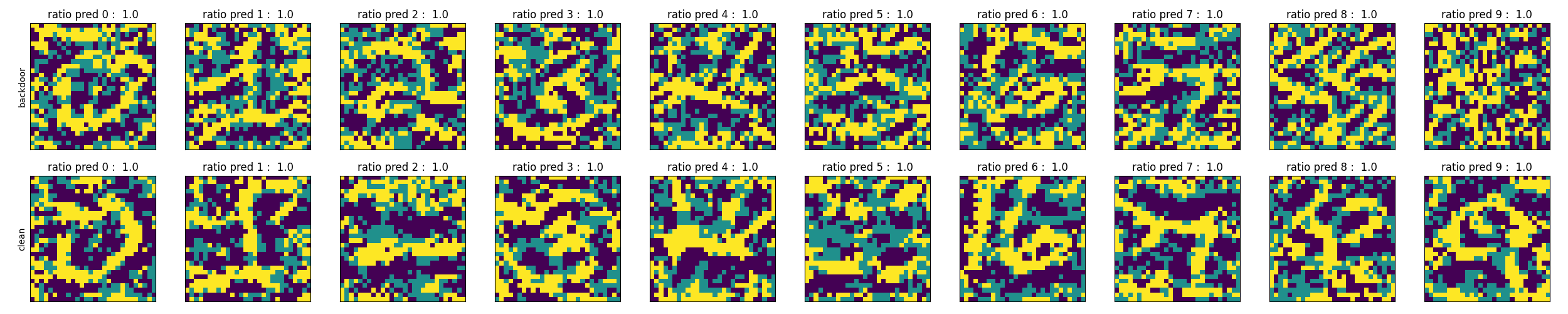}

\ \ \ \ Using data \\
\includegraphics[width=1\linewidth]{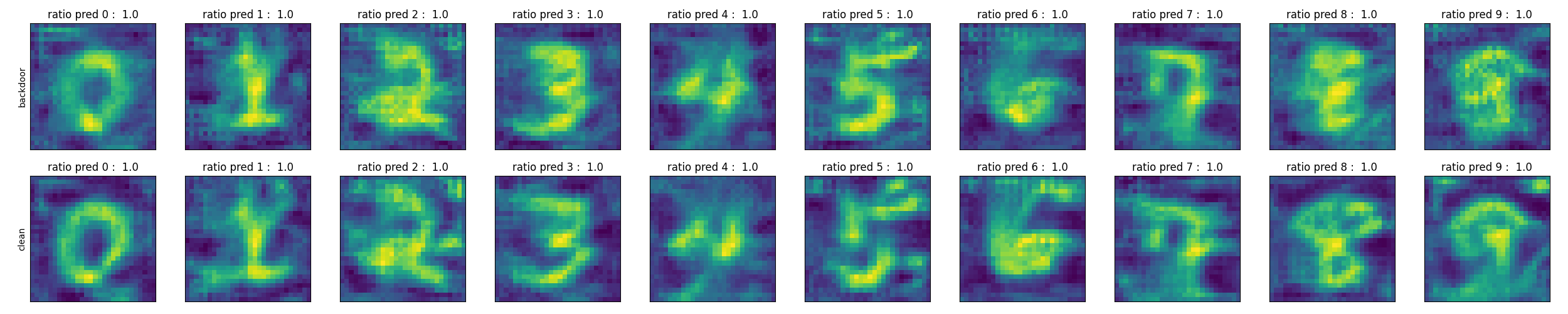}
\vspace{10pt}
\includegraphics[width=1\linewidth]{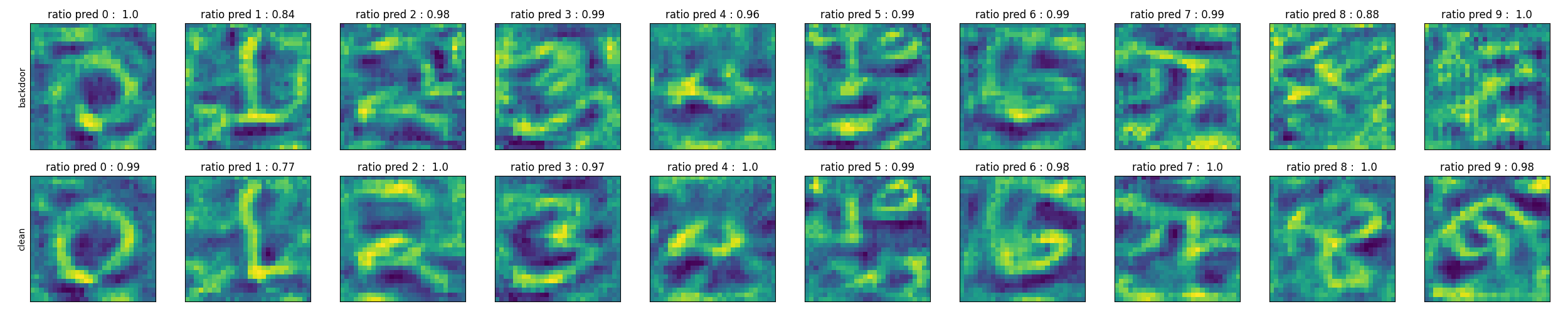}

\ \ \ \ Using noise \\
\includegraphics[width=1\linewidth]{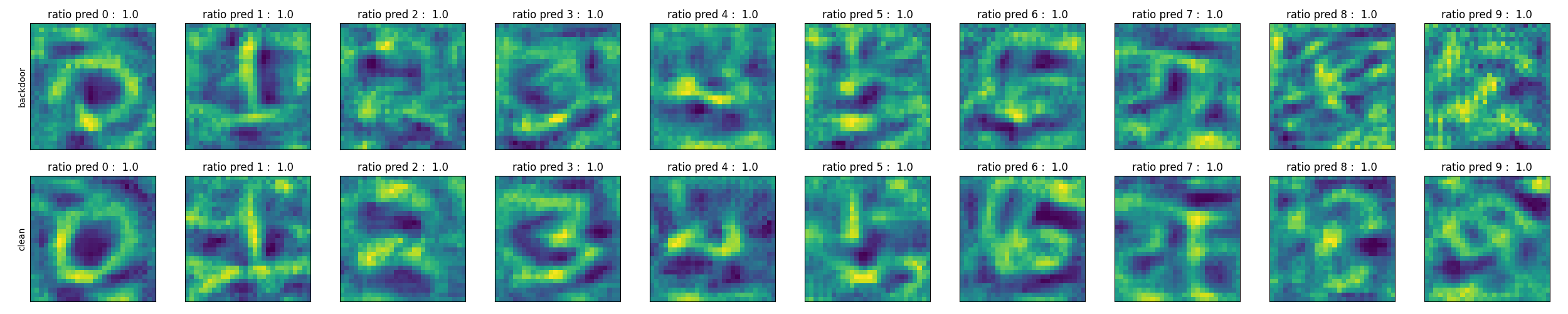}
\vspace{10pt}
\includegraphics[width=1\linewidth]{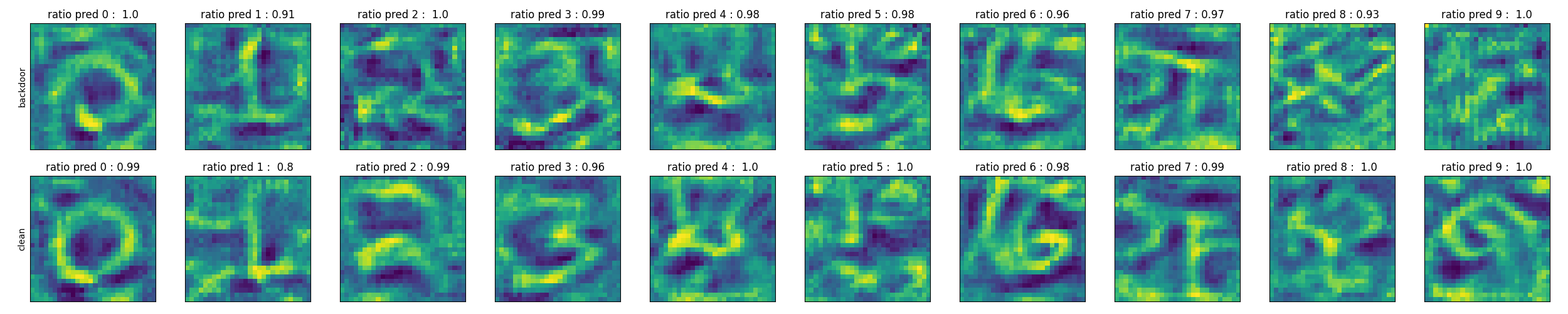}

\vspace{-10pt}
\caption{Bias maps using blank (1 sample, and 100 iterations), data, and white noise inputs (100 samples and 10 iterations) for the $8\rightarrow9$  multiplication backdoor attack. In each panel, paired rows correspond to $\epsilon$ equal to 0.1 and .5, respectively.}
\label{fig:resMinuteMult}
\end{adjustwidth}
\end{figure}

\begin{figure}[htbp]    
\vspace{-30pt}
\begin{adjustwidth}{-2cm}{-2cm}
\ \ \ \ Using blank image \\
\includegraphics[width=1\linewidth]{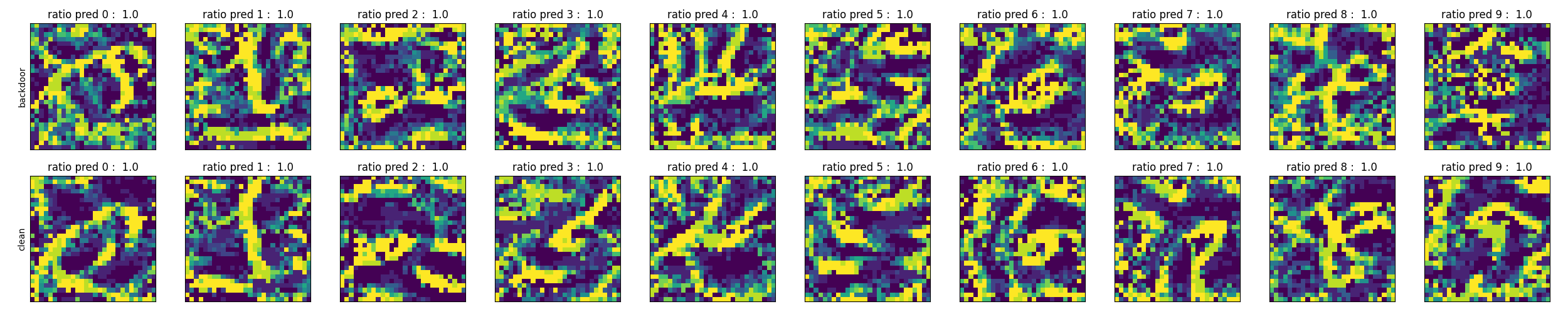}
\vspace{10pt}
\includegraphics[width=1\linewidth]{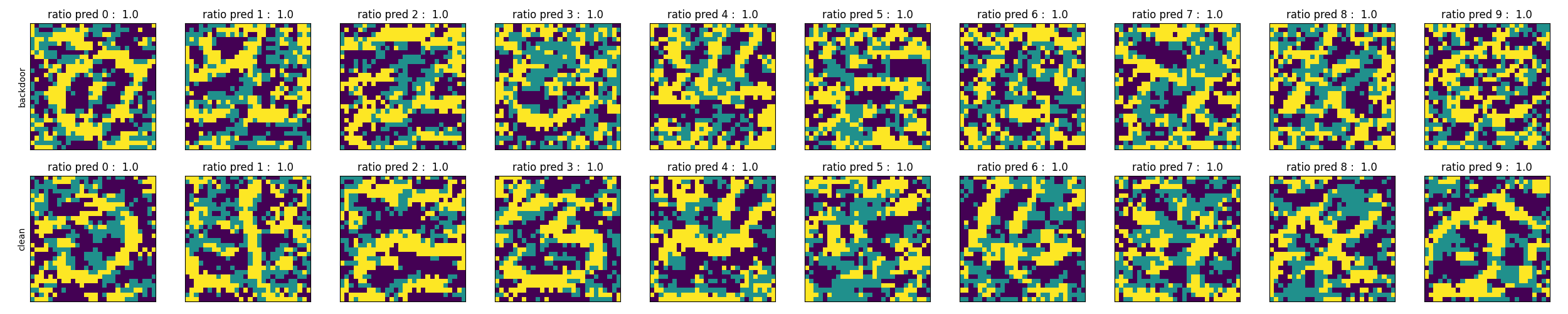}

\ \ \ \ Using data \\
\includegraphics[width=1\linewidth]{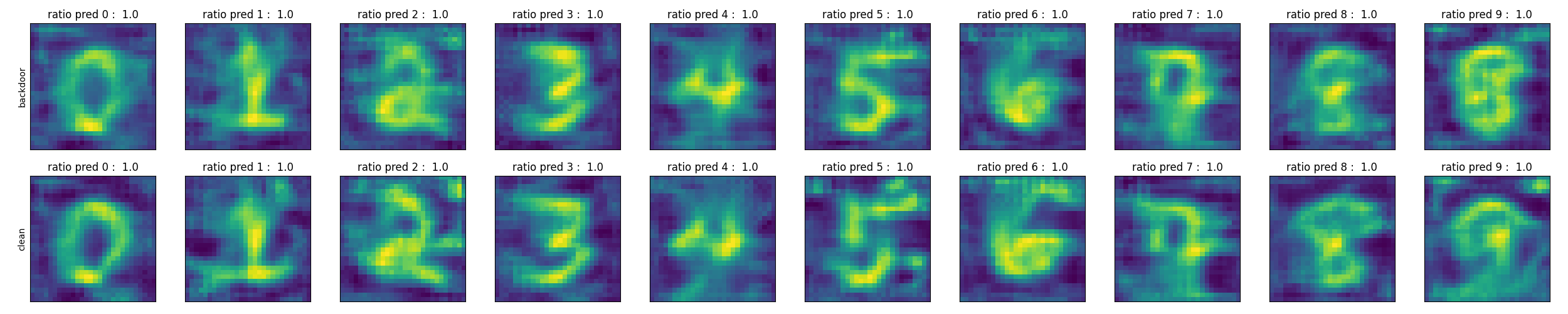}
\vspace{10pt}
\includegraphics[width=1\linewidth]{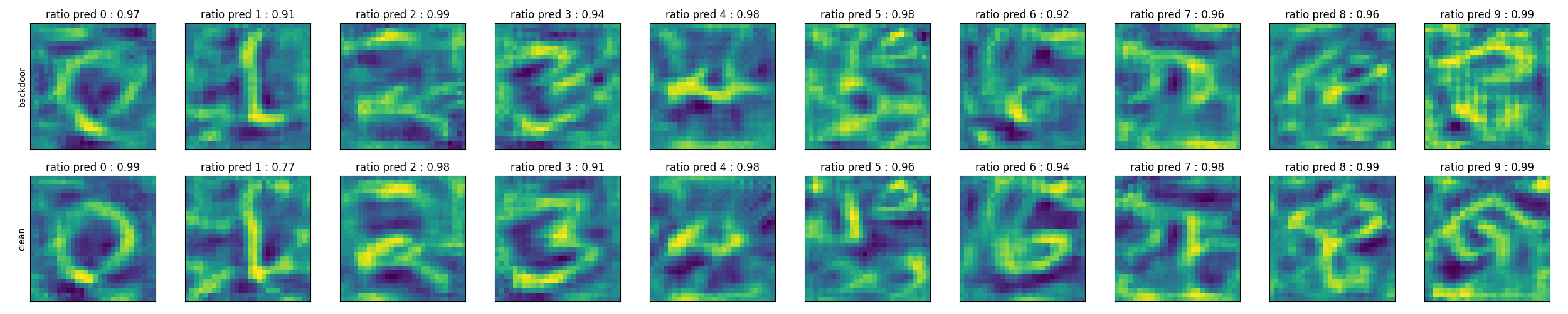}

\ \ \ \ Using noise \\
\includegraphics[width=1\linewidth]{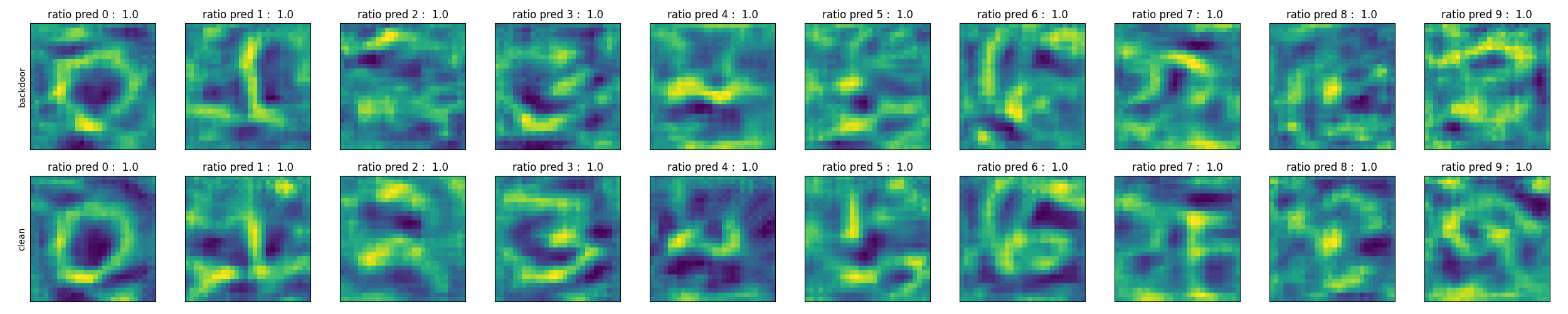}
\vspace{10pt}
\includegraphics[width=1\linewidth]{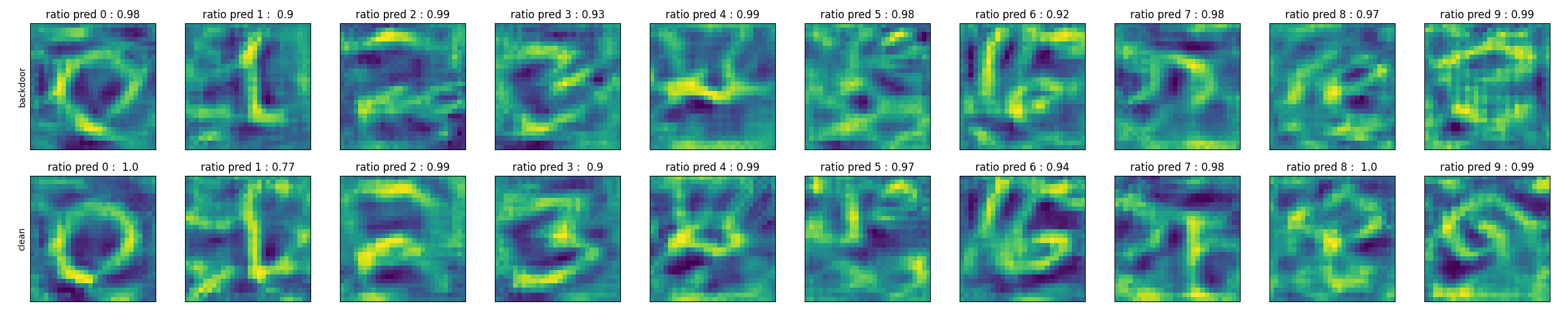}

\vspace{-10pt}
\caption{Bias maps using blank (1 sample, and 100 iterations), data, and white noise inputs (1000 samples and 10 iterations) for the $8\rightarrow9$ blending backdoor attack. In each panel, paired rows correspond to $\epsilon$ equal to 0.1 and .5, respectively.}
\label{fig:resMinuteBlend}
\end{adjustwidth}
\end{figure}

\begin{figure}
\centering
\subfigure{\includegraphics[width=.45\linewidth]{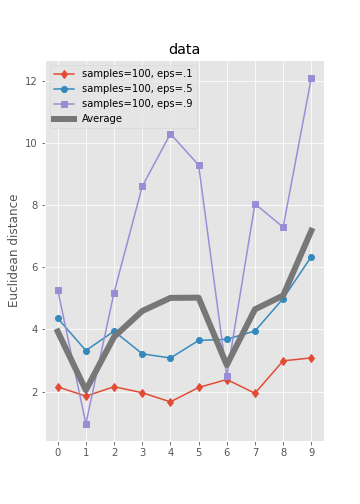}}
\vspace{-20pt}
\subfigure{\includegraphics[width=.45\linewidth]{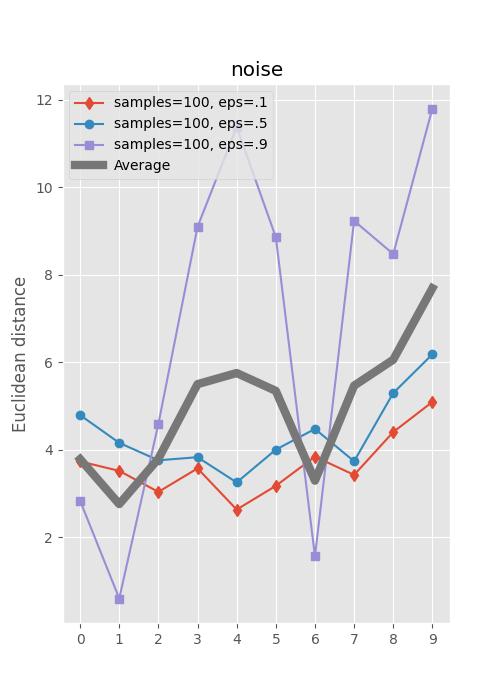}}\\
\subfigure{\includegraphics[width=.45\linewidth]{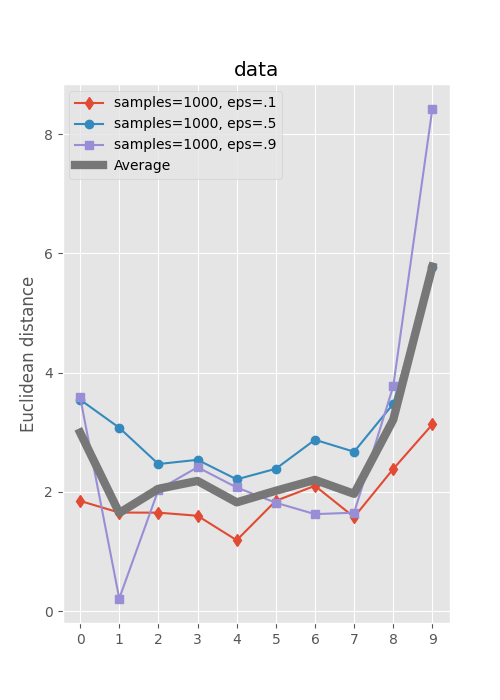}}
\subfigure{\includegraphics[width=.45\linewidth]{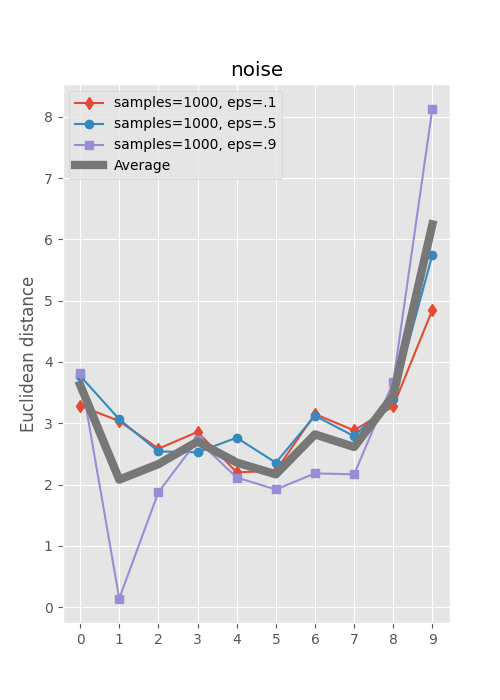}}
\caption{Quantitative performance of Algorithm I in detecting the multiplication (top) and blending (bottom) backdoor attacks (experiment three).}
\label{fig:resMinuteBar}
\end{figure}

\begin{figure}
\centering
\includegraphics[width=\linewidth]{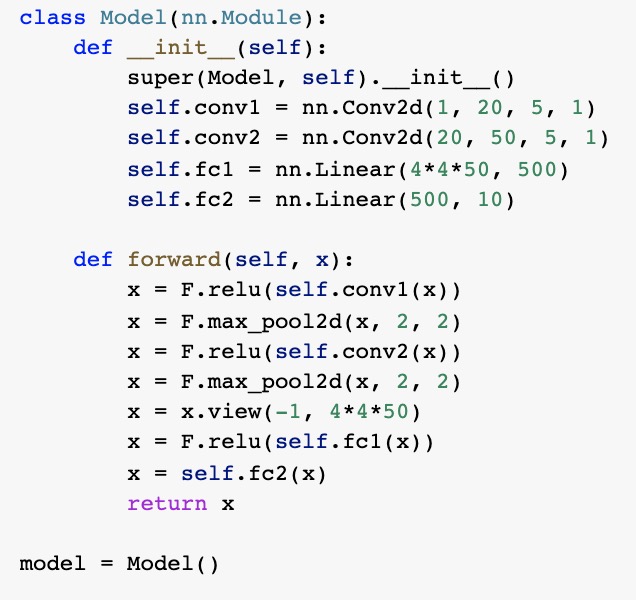}
\caption{CNN architecture used for MNIST classification.}
\label{fig:model}
\end{figure}

Qualitative results are shown in Figs.~\ref{fig:resMinuteMult} and~\ref{fig:resMinuteBlend}. Although not very clear, it seems there is a bigger difference over bias maps for digit 9 compared to other digits. Quantitative results in Fig.~\ref{fig:resMinuteBar} corroborate this observation. Admittedly though, I think this needs further investigation.

\section{Discussion}
I proposed a simple method to examine a model that might have been attacked. This method can be considered as a system identification method. With no information about a model other than output labels, one strategy is to feed white noise input to the system and analyze the statistics of noise in different categories. This is what we did in~\cite{borji2019white}. The drawback however is the demand for a large number of queries. Here, assuming the model is white-box, I was able to lower the sample complexity dramatically. For example, for 10 categories and 1000 samples per category, 10,000 samples are needed in total, whereas in~\cite{borji2019white} we used samples in the order of millions. Also, the introduced methods are much more efficient.

The proposed method differs from the existing backdoor attack detection works which often rely on statistical analysis of the poisoned training dataset (\eg~\cite{steinhardt2017certified,turner2018clean}) or the neural activations in different layers (\eg~\cite{chen2018detecting}). Here instead, I derived the biases of the networks. It is possible to improve upon the presented results by performing analysis over the individual adversarial examples in addition to the mean maps.



The preliminary results here without fine tuning or regularization (as is done in these types of works \eg DeepDream) are promising. However, further effort is needed to scale up this method over datasets containing natural scenes such as CIFAR or ImageNet.

Lastly, I wonder whether these methods can be used in neuroscience to understand feature selectivity of neurons or perceptual biases in humans, similar to~\cite{bashivan2019neural}. 
The current work also relates to neuropsychology and category learning research.

\vspace{10pt}
\noindent {\bf Acknowledgement}. I would like to thank Google for making the Colaboratory platform available.

{\small
\bibliographystyle{unsrtnat}
\bibliography{refs}
}

\clearpage

\clearpage


\clearpage

\clearpage

\clearpage

\end{document}